%% file: main.tex
\documentclass[final]{cvpr}
\usepackage{times}
\usepackage{epsfig}
\usepackage{graphicx}
\usepackage{amsmath}
\usepackage{amssymb}
\usepackage{subfigure}
\usepackage{multirow}
\usepackage{comment}
\usepackage[dvipsnames]{xcolor}

\usepackage[pagebackref=true,breaklinks=true,colorlinks,bookmarks=false]{hyperref}

\def\cvprPaperID{3036} 
\def\confYear{CVPR 2021}

\begin{document}

\title{Video Rescaling Networks with Joint Optimization Strategies \\ for Downscaling and Upscaling}


\newcommand*{\affaddr}[1]{#1} 
\newcommand*{\affmark}[1][*]{\textsuperscript{#1}}
\newcommand*{\email}[1]{\texttt{#1}}
\makeatletter
\newcommand{\printfnsymbol}[1]{%
  \textsuperscript{\@fnsymbol{#1}}%
}
\makeatother
%
\author{%
Yan-Cheng Huang\affmark[1,]\thanks{Both authors contributed equally to this work.} \qquad Yi-Hsin Chen\affmark[1,]\printfnsymbol{1} \qquad Cheng-You Lu\affmark[1]\\ Hui-Po Wang\affmark[2] \qquad Wen-Hsiao Peng\affmark[1] \qquad Ching-Chun Huang\affmark[1] \\
\affaddr{\affmark[1] National Yang Ming Chiao Tung University, Taiwan \\ \affmark[2] CISPA Helmholtz Center for Information Security}\\
\tt\small \{s0756722.iie07g, yhchen.iie07g, johnny305.cs04\}@nctu.edu.tw \\ \tt\small hui.wang@cispa.saarland, \{wpeng, chingchun\}@cs.nctu.edu.tw
}

\maketitle


\begin{abstract}
This paper addresses the video rescaling task, which arises from the needs of adapting the video spatial resolution to suit individual viewing devices. We aim to jointly optimize video downscaling and upscaling as a combined task. Most recent studies focus on image-based solutions, which do not consider temporal information. We present two joint optimization approaches based on invertible neural networks with coupling layers. Our Long Short-Term Memory Video Rescaling Network (LSTM-VRN) leverages temporal information in the low-resolution video to form an explicit prediction of the missing high-frequency information for upscaling. Our Multi-input Multi-output Video Rescaling Network (MIMO-VRN) proposes a new strategy for downscaling and upscaling a group of video frames simultaneously. Not only do they outperform the image-based invertible model in terms of quantitative and qualitative results, but also show much improved upscaling quality than the video rescaling methods without joint optimization. To our best knowledge, this work is the first attempt at the joint optimization of video downscaling and upscaling.  
\end{abstract}

\input{intro.tex}

\input{related.tex}

\input{method.tex}
\input{experiments.tex}

\input{conclusion.tex}
\paragraph{\textbf{Acknowledgements.}} {This work is supported by Qualcomm technologies, Inc. (NAT-439543), Ministry of Science and Technology, Taiwan (109-2634-F-009-020) and National Center for High-performance Computing, Taiwan.}

{\small
\bibliographystyle{ieee_fullname}
\bibliography{egbib}
}
\clearpage

\appendix
\title{\noindent\LARGE\textbf{Appendix}}
\maketitle
\vspace{4mm}
\input{appendix.tex}
\end{document}

%% file: intro.tex
\section{Introduction}
\label{sec:intro}

With the increasing popularity of video capturing devices, a tremendous amount of high-resolution (HR) videos are shot every day. These HR videos are often downscaled to save storage space and streaming bandwidth, or to fit screens with lower resolutions. It is also common that the downscaled videos need to be upscaled for display on HR monitors~\cite{kim2018task, li2018learning, sun2020learned, chen2020hrnet, xiao2020invertible}. 


\begin{figure}[t]
\centering
\subfigure[SISO-down-SISO-up]{
\label{fig:fig1-a}
\centering
\includegraphics[width=.46\textwidth, trim= 0 755 0 0, clip]{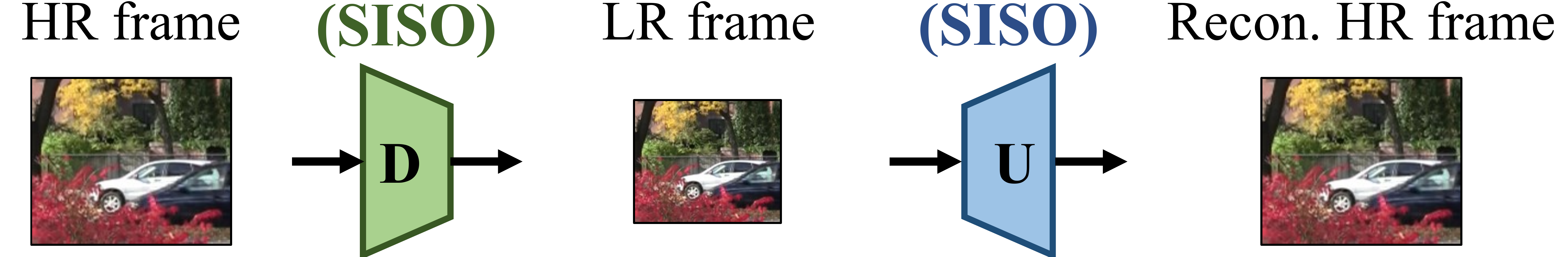}
}
\subfigure[SISO-down-MISO-up]{
\label{fig:fig1-b}
\centering
\includegraphics[width=.46\textwidth, trim= 0 340 0 340, clip]{figures/MIMO_v8.pdf}
}
\subfigure[MIMO-down-MIMO-up (the proposed method)]{
\label{fig:fig1-c}
\centering
\includegraphics[width=.46\textwidth, trim= 0 0 0 713, clip]{figures/MIMO_v8.pdf}
}
\caption{Comparison of video rescaling frameworks according to the downscaling and upscaling strategies: (a) single-input single-output (SISO) for both operations, (b) SISO for downscaling and multi-input single-output (MISO) for upscaling, and (c) multi-input multi-output (MIMO) for both operations (the proposed method).}
\label{fig:fig1}
\end{figure}

In this paper, we address the joint optimization of video downscaling and upscaling as a combined task, which is referred to as \textit{video rescaling}. This task involves downscaling an HR video into a low-resolution (LR) one, followed by upscaling the resulting LR video back to HR. Our aim is to optimize the HR reconstruction quality while regularizing the LR video to offer comparable visual quality to the bicubic-downscaled video for human perception. It is to be noted that the rescaling task differs from the super-resolution task; at inference time, the former has access to the HR video while the latter has no such information.

One straightforward solution to video rescaling is to downscale an HR video by predefined kernels and upscale the LR video with super-resolution methods~\cite{8100101, lim2017enhanced, zhang2018image, wang2018esrgan, dai2019second, guo2020closed,caballero2017real, tao2017detail, sajjadi2018frame, jo2018deep, wang2019edvr, yi2019progressive, isobe2020vide, li2020mucan, isobe2020video}. With this solution, the downscaling is operated independently of the upscaling although the upscaling can be optimized for the chosen downscaling kernels. The commonly used downscaling (e.g.~bicubic) kernels suffer from losing the high-frequency information~\cite{shannon1949communication} inherent in the HR video, thus creating a many-to-one mapping between the HR and LR videos. Reconstructing the HR video by upscaling its LR representation becomes an ill-posed problem. The independently-operated downscaling misses the opportunity of optimizing the downscaled video to mitigate the ill-posedness.

The idea of jointly optimizing downscaling and upscaling was first proposed for image rescaling~\cite{kim2018task, li2018learning, sun2020learned, chen2020hrnet}. It adds a new dimension of thinking to the studies of learning specifically to upscale for a given downscaling method \cite{8100101, lim2017enhanced, zhang2018image, wang2018esrgan, dai2019second, guo2020closed}. Recognizing the reciprocality of the downscaling and upscaling operations, 
IRN~\cite{xiao2020invertible} recently introduced a coupling layer-based invertible model, which shows much improved HR reconstruction quality than the non-invertible models.

These jointly optimized image-based solutions (Fig.~\ref{fig:fig1-a}) are not ideal for video rescaling. For example, a large number of prior works~\cite{caballero2017real, tao2017detail, sajjadi2018frame, jo2018deep, wang2019edvr, yi2019progressive, isobe2020vide, li2020mucan, isobe2020video} for video upscaling have adopted the Multi-Input Single-Output (MISO) strategy to reconstruct one HR frame from multiple LR frames and/or previously reconstructed HR frames (Fig.~\ref{fig:fig1-b}). They demonstrate the potential for recovering the missing high-frequency component of a video frame from temporal information. However, image-based solutions do not consider temporal information. In addition, two issues remain widely open as (1) how video downscaling and upscaling could be jointly optimized and (2) how temporal information could be utilized in the joint optimization framework to benefit both operations.

In this paper, we present two joint optimization approaches to video rescaling: Long Short-Term Memory Video Rescaling Network (LSTM-VRN) and Multi-Input Multi-Output Video Rescaling Network (MIMO-VRN). The LSTM-VRN downscales an HR video frame-by-frame using a similar coupling architecture to \cite{xiao2020invertible}, but fuses multiple downscaled LR video frames via LSTM to estimate the missing high-frequency component of an LR video frame for upscaling (Fig.~\ref{fig:fig1-b}). LSTM-VRN shares similar downscaling and upscaling strategies to the traditional video rescaling framework. In contrast, our MIMO-VRN introduces a completely new paradigm by adopting the MIMO strategy for both video downscaling and upscaling (Fig.~\ref{fig:fig1-c}). We develop a group-of-frames-based (GoF) coupling architecture that downscales multiple HR video frames simultaneously, with their high-frequency components being estimated also simultaneously in the upscaling process. 
Our contributions include the following:
\begin{itemize}

\item To the best of our knowledge, this work is the first attempt at jointly optimizing video downscaling and upscaling with invertible coupling architectures. 

\item Our LSTM-VRN and MIMO-VRN outperform the image-based invertible model~\cite{xiao2020invertible}, showing significantly improved HR reconstruction quality and offering LR videos comparable to the bicubic-downscaled video in terms of visual quality.  

\item Our MIMO-VRN is the first scheme to introduce the MIMO strategy for video upscaling and downscaling, achieving the state-of-the-art performance.

\end{itemize}

%% file: related.tex
\section{Related Work}
\label{sec:related}
\begin{figure}[t]
\begin{center}
\includegraphics[width=.47\textwidth]{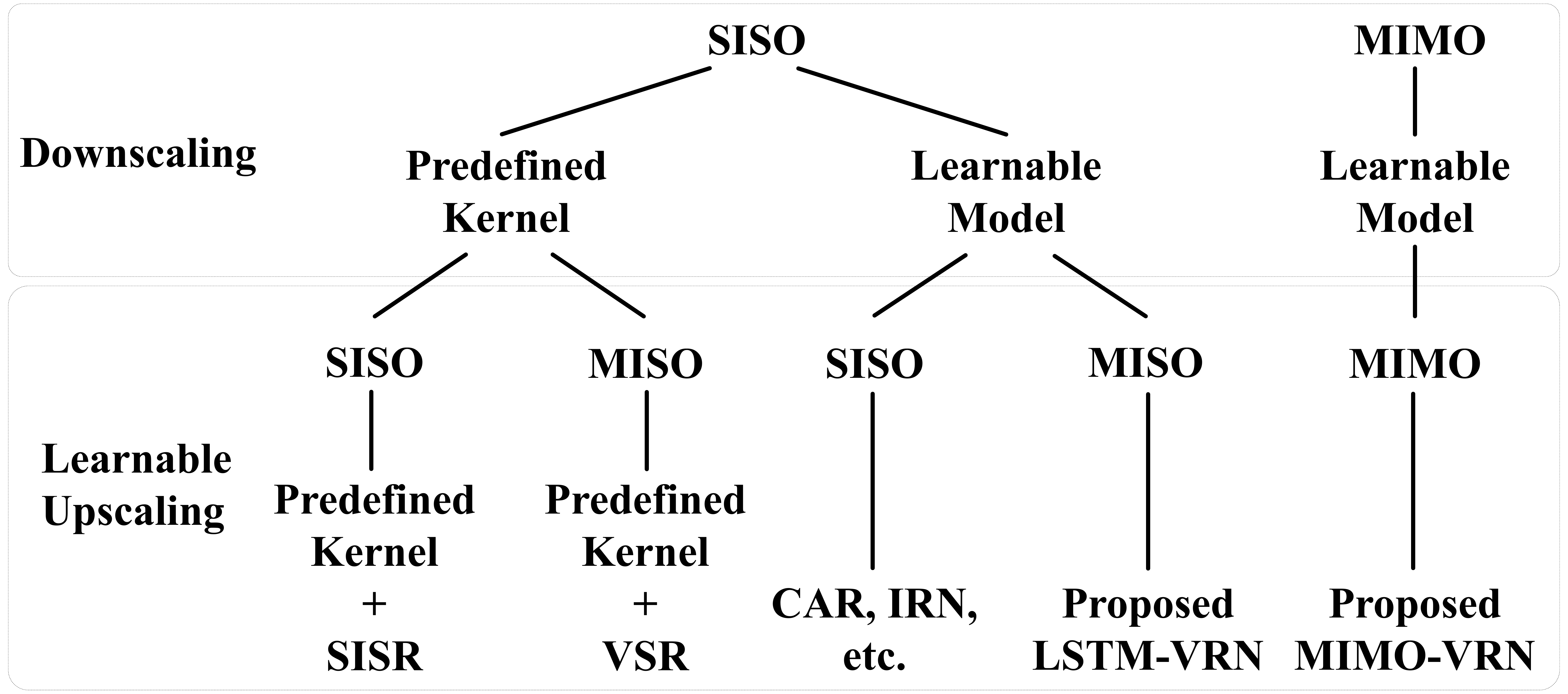}
\end{center}
\caption{Taxonomy of the prior works on image/video rescaling. The SISO, MISO and MIMO indicate the strategies (i.e.~the input/output format) for downscaling and upscaling. SISR and VSR stand for single image super-resolution and video super-resolution, respectively. CAR \cite{sun2020learned} and IRN \cite{xiao2020invertible} are joint optimization schemes for image rescaling.}
\label{fig:tree}
\end{figure}
This section surveys video rescaling methods, with a particular focus on their downscaling and upscaling strategies. We regard the image-based rescaling methods as possible solutions for video rescaling. Fig.~\ref{fig:tree} is a taxonomy of these prior works. 

\subsection{Upscaling with Predefined Downscaling}
The traditional image super-resolution~\cite{8100101, lim2017enhanced, zhang2018image, wang2018esrgan, dai2019second, guo2020closed} or video super-resolution~\cite{caballero2017real, tao2017detail, sajjadi2018frame, jo2018deep, wang2019edvr, yi2019progressive, isobe2020vide, li2020mucan, isobe2020video} methods are candidate solutions to video upscaling. The former is naturally a single-input single-output (SISO) upscaling strategy, which generates one HR image from one LR image. The latter usually involves more than one LR video frame in the upscaling process, i.e.~the MISO upscaling strategy, in order to leverage temporal information for better HR reconstruction quality. Most of the approaches in this category adopt a SISO downscaling strategy with a pre-defined kernel (e.g.~bicubic) chosen independently of the upscaling process. Therefore, they are unable to adapt the downscaled images/videos to the upscaling.

\subsection{Upscaling with Jointly Learned Downscaling}
To mitigate the ill-posedness of the image upscaling task, some works learn upscaling and downscaling jointly by encoder-decoder architectures~\cite{kim2018task, li2018learning, sun2020learned, chen2020hrnet}. They turn the fixed downscaling method into a learnable model in order to adapt the LR image to the upscaling process that is learned jointly. The training objective usually requires the LR image to be also suitable for human perception. Recently, IRN~\cite{xiao2020invertible} introduces an invertible model \cite{DBLP:journals/corr/DinhKB14, DBLP:conf/iclr/DinhSB17,  kingma2018glow} to this joint optimization task. It is able to perform image downscaling and upscaling by the same set of neural networks configured in the reciprocal manner. It provides a means to model explicitly the missing high-frequency information due to downscaling by a Gaussian noise.

\label{sec:inn}
\begin{figure}[t]
\begin{center}
\includegraphics[width=.45\textwidth]{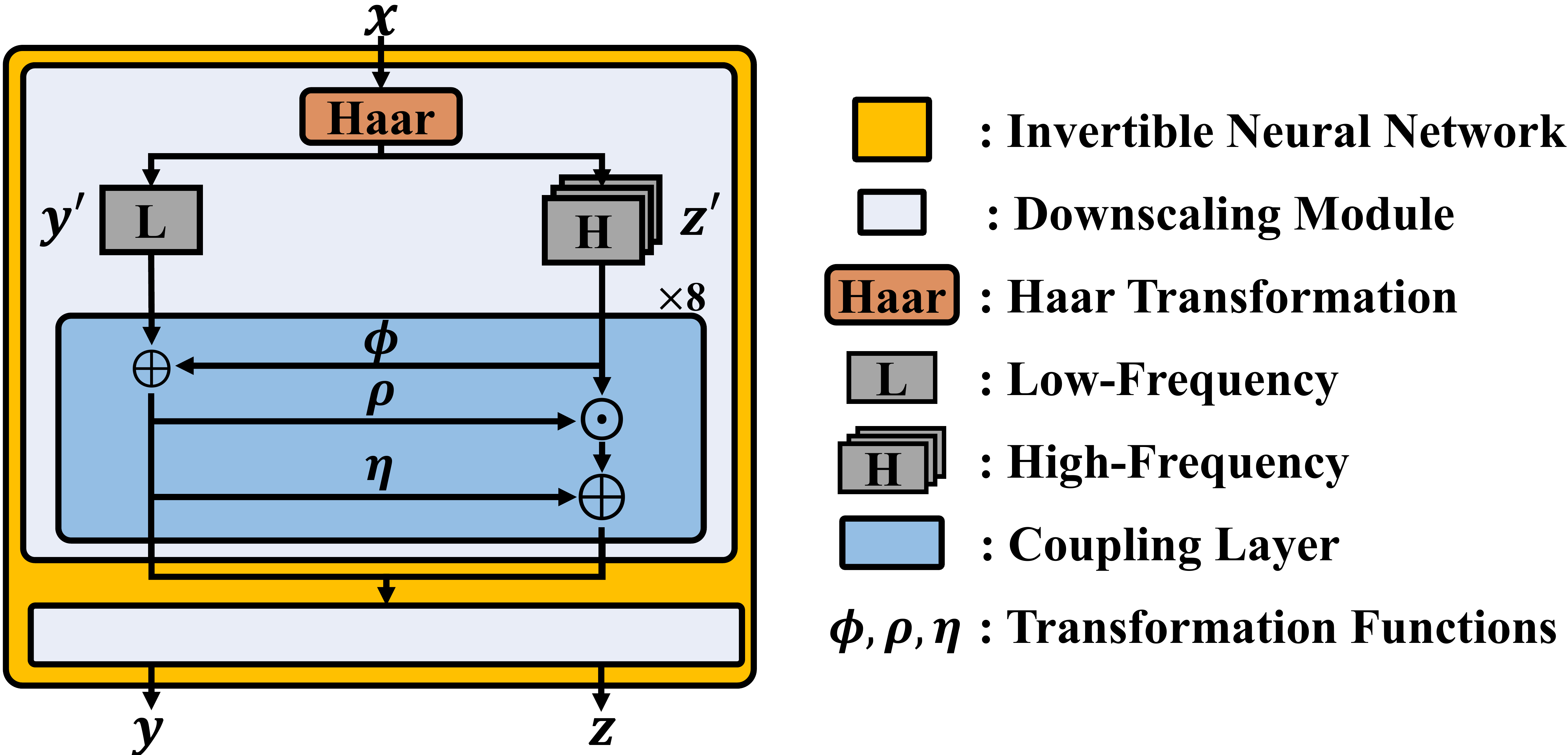}
\end{center}
\vspace{-2mm}
\caption{The detailed downscaling operation of IRN~\cite{xiao2020invertible}. The model performs $\times4$ downscaling with two downscaling modules, each of which comprises a 2-D Haar transform and eight coupling layers. Each downscaling module halves the horizontal and vertical resolutions of the input image.
}
\label{fig:INN}
\end{figure}

\subsection{Invertible Rescaling Network}
\label{sec:irn}
IRN~\cite{xiao2020invertible} is an invertible model designed specifically for image rescaling.
The forward model of IRN comprises a 2-D Haar transform and eight coupling layers~\cite{DBLP:journals/corr/DinhKB14, DBLP:conf/iclr/DinhSB17, kingma2018glow}, as shown in Fig.~\ref{fig:INN}. By applying the 2-D Haar transform, an input image $x \in \mathbb{R}^{C\times H \times W}$ is first decomposed into one low-frequency band $y' \in \mathbb{R}^{C \times \frac{H}{2} \times \frac{W}{2}}$ and three other high-frequency bands $z' \in \mathbb{R}^{3C \times \frac{H}{2} \times \frac{W}{2}}$. These two components $y',z'$ are subsequently processed via the coupling layers in a way that the output $y$ becomes a visually-pleasing LR image and the $z$ encodes the complementary high-frequency information inherent in the input HR image $x$. In theory, the inverse coupling layers can recover $x$ losslessly from $y$ and $z$ because the model is invertible. In practice, $z$ is unavailable for upscaling at inference time. The training of IRN requires $z$ to follow a Gaussian distribution so that at inference time, a Gaussian sample $\hat{z}$ can be drawn as a substitute for the missing high-frequency component.

Although IRN achieves superior results on the image rescaling task, it is not optimal for video rescaling. Essentially, IRN is an image-based method. This work presents the first attempt at jointly optimizing video downscaling and upscaling with an invertible coupling architecture (Fig.~\ref{fig:INN}).

%% file: method.tex
\section{Proposed Method}
\label{sec:method}

\begin{figure*}[t]
\centering
\subfigure[LSTM-VRN]{
\centering
\includegraphics[scale=0.155,trim=0 0 1620 0,clip]{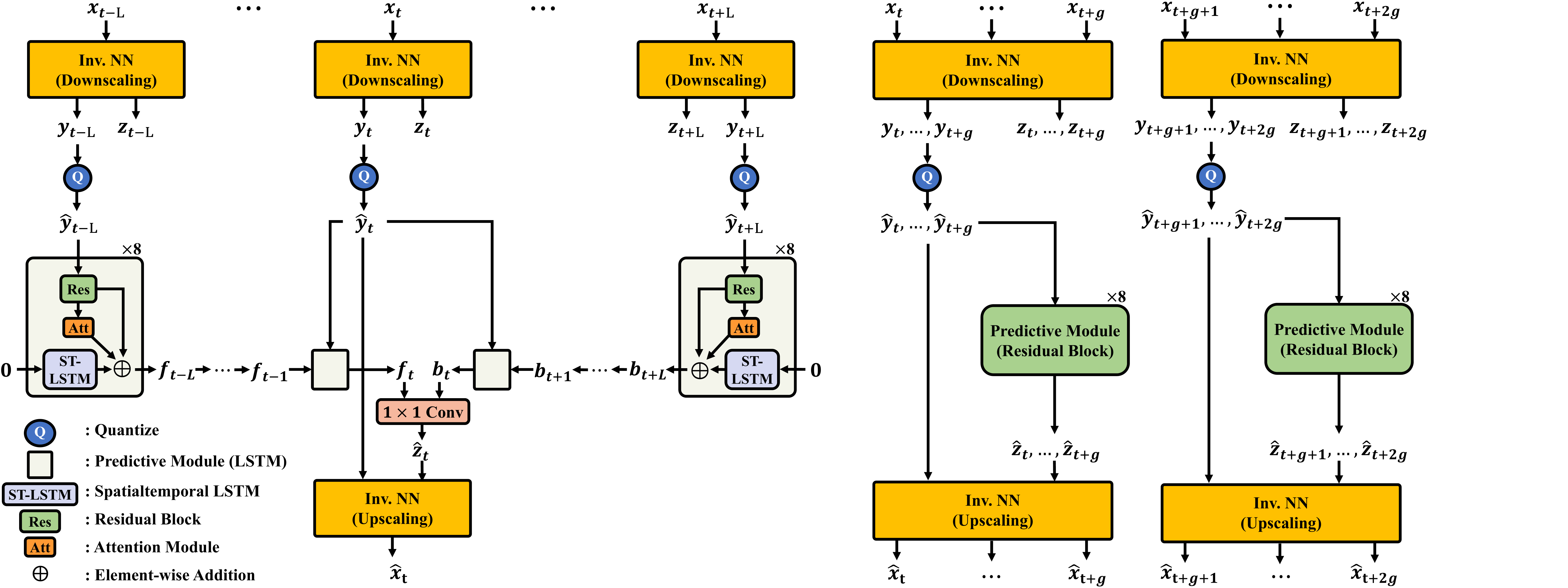}
\label{fig:LSTM}
}
\subfigure[MIMO-VRN]{
\centering
\includegraphics[scale=0.155,trim=1850 0 300 0,clip]{figures/LSTM_GOP_v8.pdf}
\label{fig:GoP}
}
\caption{Overview of the proposed LSTM-VRN and MIMO-VRN for video rescaling. Both schemes involve an invertible network with coupling layers for video downscaling and upscaling. In part (a), LSTM-VRN downscales every video frame $x_t$ independently and forms a prediction $\hat{z}_t$ of the high-frequency component $z_t$ from the LR video frames $\{\hat{y}_i\}_{i=t-L}^{t+L}$ by a bi-directional LSTM that operates in a sliding window manner. In part (b), MIMO-VRN downscales a group of HR video frames $\{x_i\}_{i=t}^{t+g}$ into the LR video frames $\{\hat{y}_i\}_{i=t}^{t+g}$ simultaneously. The upscaling is also done on a group-by-group basis, with the high-frequency components $\{z_i\}_{i=t}^{t+g}$ estimated from the $\{\hat{y}_i\}_{i=t}^{t+g}$ by a predictive module.}
\label{fig:Architecture}
\end{figure*}

Given an HR video composed of $N$ video frames $\{x_t\}_{t=1}^N$, where $x_t \in \mathbb{R}^{C \times H \times W}$, the video rescaling task involves (1) downscaling every video frame $x_t$ to its LR counterpart $y_t\in \mathbb{R}^{C \times \frac{H}{4} \times \frac{W}{4}}$, where the quantized version $\hat{y}_t$ of which forms collectively an LR video $\{\hat{y}_t\}_{t=1}^N$, and (2) upscaling the LR video to arrive at the reconstructed HR video $\{\hat{x}_t\}_{t=1}^N$. Unlike most video super-resolution tasks, which focus primarily on learning upscaling for a given downscaling method, this work optimizes jointly the downscaling and upscaling as a combined task. It has been shown in many traditional video super-resolution works~~\cite{caballero2017real, tao2017detail, sajjadi2018frame, haris2019recurrent, tian2020tdan, wang2019edvr, jo2018deep, yi2019progressive, isobe2020video, li2020mucan, isobe2020vide} that the extra temporal information in videos allows the lost high-frequency component of a video frame due to downscaling to be recovered to some extent. This work makes the first attempt to explore how such temporal information could assist downscaling in producing an LR video that can be upscaled to offer better super-resolution quality in an end-to-end fashion. In a sense, our focus is on both downscaling and upscaling. The objective is to minimize the distortion between $\{\hat{x}_t\}_{t=1}^N$ and $\{x_t\}_{t=1}^N$ in such a combined task while the LR video $\{\hat{y}_t\}_{t=1}^N$ is regularized to offer comparable visual quality to the bicubic-downscaled video for human perception. It is to be noted that the LR video is not meant to be exactly the same as the bicubic-downscaled video since doing so may not lead to the optimal downscaling and upscaling in our task.   

The reciprocality of the downscaling and upscaling operations motivates us to choose an invertible network for our task. With the superior performance of coupling layer architectures in recovering high-frequency details of LR images~\cite{xiao2020invertible}, we develop our downscaling and upscaling networks, especially for video, using a similar invertible architecture (Sec.~\ref{sec:irn}) as the basic building block. 

We propose two approaches, LSTM-VRN and MIMO-VRN, to configure or extend these building blocks for joint learning of video downscaling and upscaling. Their overall architectures are depicted in Fig.~\ref{fig:Architecture}, with detailed operations given in the following sections. 



\subsection{LSTM-based Video Rescaling Network}

Like most video super-resolution techniques, the LSTM-VRN (Fig.~\ref{fig:LSTM}) adopts the SISO strategy to downscale HR video frames $\{x_t\}_{t=1}^N$ individually to their LR ones $\{\hat{y}_t\}_{t=1}^N$ by the forward model of the invertible network. The operation is followed by the MISO-based upscaling, which departs from the idea of drawing an input-agnostic Gaussian noise $\cite{xiao2020invertible}$ for complementary high-frequency information. Specifically, we fuse the current LR frame ${\hat{y}_t}$ and its neighbouring frames $\{\hat{y}_{t-i},\hat{y}_{t+i}\}_{i=1}^L$ by a LSTM-based predictive module to form an estimate $\hat{z}_t$ of the missing high-frequency component $z_t$ at inference time. The resulting $\hat{z}_t$ is fed to the inverse model together with the ${\hat{y}_t}$ for reconstructing the HR video frame $\hat{x}_t$. The fact that $z_t$ needs to be estimated from multiple LR frames $\{\hat{y}_i\}_{i=t-L}^{t+L}$ determines what information should remain in the LR video to facilitate the prediction. This connects the upscaling process tightly to the downscaling process, stressing the importance of their joint optimization. In addition, we rely on the inter-branch pathways of the coupling layer in the forward model to correlate $z_t$ and $y_t$ in such a way that $z_t$ could be better predicted from $\hat{y}_t$ and its neighbors $\{\hat{y}_{t-i},\hat{y}_{t+i}\}_{i=1}^L$.


The predictive module plays a key role in fusing information from $\hat{y}_t$ and $\{\hat{y}_{t-i},\hat{y}_{t+i}\}_{i=1}^L$. We incorporate Spatiotemporal-LSTM (ST-LSTM)~\cite{wang2017predrnn} for propagating temporal information in both forward and backward directions, in view of its recent success in video extrapolation tasks. Eq.~\eqref{equ:lstm} details the forward mode of the predictive module for time instance $t$:
\begin{equation}
\label{equ:lstm}
\begin{aligned}
 h^f_t &= ST\mbox{-}LSTM(f_{t-1}, h^f_{t-1}) \\
 h^y_t &= ResidualBlock(\hat{y}_t)\\
 a_t &= \sigma (W \otimes h^y_t)\\
 f_t &= (1-a_t) \odot h^f_t + a_t \odot h^y_t
\end{aligned}
\end{equation}
where $\sigma$ is a sigmoid function, $\otimes$ is the standard convolution, and $\odot$ is Hadamard product. Note that an attention signal $a_t$ guided by the the current LR frame $\hat{y}_t$ combines the temporally-propagated hidden information $h^f_t$ and the features $h^y_t$ of $\hat{y}_t$ to yield the output $f_t$. As Fig.~\ref{fig:LSTM} shows, the forward propagated $f_t$ is further combined with the backward propagated $b_t$ to predict $\hat{z}_t$ through a 1x1 convolution.






For upscaling every LR video frame $\hat{y}_{t}$, the proposed predictive module works in a sliding-window manner with a window size of $2L+1$. That is, the forward (respectively, backward) ST-LSTM always starts with a reset state 0 when accepting the input $\hat{y}_{t-L}$ (respectively, $\hat{y}_{t+L}$). This design choice is out of generalization and buffering considerations. We avoid running a long ST-LSTM at inference time because the training videos are rather short. Moreover, the backward ST-LSTM introduces delay and buffering requirements.    


Finally, we note in passing that LSTM-VRN exploits temporal information across LR video frames only for upscaling while its downscaling is still a SISO-based scheme, which does not take advantage of temporal information in HR video frames for downscaling. 


\subsection{MIMO-based Video Rescaling Network}

Our MIMO-VRN (Fig.~\ref{fig:GoP}) is a new attempt that adopts a MIMO strategy for both upscaling and downscaling, making explicit use of temporal information in these operations. Here, we propose a new basic processing unit, called Group of Frames (GoF). To begin with, the HR input video $\{x_t\}_{t=1}^N$ is decomposed into non-overlapping groups of frames, with each group including $g$ frames, namely $\{x_t\}_{t=1}^g, \{x_t\}_{t=g+1}^{2g},\ldots$. The downscaling proceeds on a group-by-group basis; each GoF is downscaled independently of each other. Within a GoF, every HR video frame $x_t \in \mathbb{R}^{C \times H \times W}$ is first transformed individually using 2-D Haar Wavelet, to arrive at its low-frequency $y'_t \in \mathbb{R}^{C \times \frac{H}{2} \times \frac{W}{2}}$ and high-frequency $z'_t \in \mathbb{R}^{3C \times \frac{H}{2} \times \frac{W}{2}}$ components. We then group the low-frequency components $\{y'_i\}_{i=t}^{t+g}$ in a GoF as one group-type input $\mathcal{Y}'_t \in \mathbb{R}^{gC \times \frac{H}{2} \times \frac{W}{2}}$ to the coupling layers (i.e.~replacing $y'$ in Fig.~\ref{fig:INN} with $\mathcal{Y}'_t$) and the remaining high-frequency components $\{z'_i\}_{i=t}^{t+g}$ as the other group-type input $\mathcal{Z}'_t \in \mathbb{R}^{3gC \times \frac{H}{2} \times \frac{W}{2}}$ (i.e.~replacing $z'$ in Fig.~\ref{fig:INN} with $\mathcal{Z}'_t$). Because each group-type input contains information from multiple video frames, the coupling layers are able to utilize temporal information inherent in one group-type input to update the other. With two downscaling modules, the results are a group of quantized LR frames $\hat{\mathcal{Y}}_t=\{\hat{y}_i\}_{i=t}^{t+g}$ and the group high-frequency component $\mathcal{Z}_t=\{z_i\}_{i=t}^{t+g}$. It is worth noting that due to the nature of group-based coupling, there is no one-to-one correspondence between the signals in $\hat{\mathcal{Y}}_t\in \mathbb{R}^{gC \times \frac{H}{4} \times \frac{W}{4}}$ and  $\mathcal{Z}_t \in \mathbb{R}^{3gC \times \frac{H}{4} \times \frac{W}{4}}$. 

The upscaling proceeds also on a group-by-group basis, with the group size $g$ and the group formation fully aligned with those used for downscaling. As depicted in Fig.~\ref{fig:GoP}, we employ a residual block-based predictive module to form a prediction of the missing high-frequency components $\{z_i\}_{i=t}^{t+g}$ from the corresponding group of LR frames $\{\hat{y}_i\}_{i=t}^{t+g}$. Similar to the notion of the group-type inputs for downscaling, the LR frames $\{\hat{y}_i\}_{i=t}^{t+g}$ and the estimated high-frequency components $\{\hat{z}_i\}_{i=t}^{t+g}$ comprise respectively the two group-type inputs $\hat{\mathcal{Y}}_t$ and $\hat{\mathcal{Z}}_t$ to the invertible network operated in inverse mode. With this MIMO-based upscaling, a group of HR frames $\{\hat{x}_i\}_{i=t}^{t+g}$ are reconstructed simultaneously.



\subsection{Training Objectives}
\noindent\textbf{LSTM-VRN.}
The training of LSTM-VRN involves two loss functions to reflect our objectives. First, to ensure that the LR video $\{\hat{y}_t\}_{t=1}^N$ is visually pleasing, we follow common practice to require that $\{\hat{y}_t\}_{t=1}^N$ have similar visual quality to the bicubic-downscaled video $\{x^{bic}_t\}_{t=1}^N$; to this end, we define the LR loss as 
\begin{equation}
\mathcal{L}_{LR}=\frac{1}{N}\sum^{N}_{t=1}\|x^{bic}_t-\hat{y}_t\|^2. 
\label{equ:lr_loss}
\end{equation}
Second, to maximize the HR reconstruction quality, we minimize the Charbonnier loss ~\cite{8100101} between the original HR video $\{x_t\}_{t=1}^N$ and its reconstructed version $\{\hat{x}_t\}_{t=1}^N$ subject to downscaling and upscaling:
\begin{equation}
\mathcal{L}_{HR}=\frac{1}{N}\sum^{N}_{t=1}\sqrt{\| x_t-\hat{x}_t \|^2+\epsilon^2},
\label{equ:hr_loss}
\end{equation}
where $\epsilon$ is set to $1 \times 10^{-3}$. The total loss is $\mathcal{L}_{total} 
= \mathcal{L}_{HR} + \lambda\mathcal{L}_{LR}$, where $\lambda$ is a hyper-parameter used to trade-off between the quality of the LR and HR videos.



\noindent\textbf{MIMO-VRN.}
The training of MIMO-VRN shares the same $\mathcal{L}_{LR}$ and $\mathcal{L}_{HR}$ losses as LSTM-VRN because they have common optimization objectives. We however notice that  MIMO-VRN tends to have uneven HR reconstruction quality over video frames in a GoF (Sec.~\ref{sec:ablation_study}). To mitigate the quality fluctuation in a GoF, we additionally introduce the following center loss for MIMO-VRN:
\begin{equation}
\mathcal{L}_{center} = \frac{1}{M \times g}\sum_{m=1}^{M} \sum^{mg}_{t=(m-1)g+1} \left| \|x_t-\hat{x}_t\|^2 - c_m \right|,
\label{equ:center_loss}
\end{equation}
where $g$ is the group size, $c_m = \sum^{mg}_{t=(m-1)g+1} \|x_t-\hat{x}_t\|^2/g$ denotes the average HR reconstruction error in a GoF, and $M$ is the number of GoF's in a sequence. Eq.~\eqref{equ:center_loss} encourages the HR reconstruction error of every video frame in a GoF to approximate the average level $c_m$. 
%


%% file: experiments.tex
\section{Experimental Results}
\label{sec:experiments}
\subsection{Setup}
\noindent\textbf{Datasets.} 
For a fair comparison, we follow the common test protocol to train our models on Vimeo-90K dataset~\cite{Xue_2019}. It has 91,701 video sequences, each is 7 frames long. Among them, 64,612 sequences are for training and 7,824 are for test. Each sequence has a fixed spatial resolution of 448~$\times$~256. The performance evaluation is done on two standard test datasets, Vimeo-90K-T and Vid4~\cite{6549107}. Vid4 includes 4 video clips, each having around 40 frames. 

\input{tables/vid4_results}

\noindent\textbf{Implementation and Training Details.}
Our proposed models adopt the settings from IRN~\cite{xiao2020invertible}, which consists of two downscaling modules (Fig.~\ref{fig:INN}). Each module is composed of one 2-D Haar transform and eight coupling layers. Both LSTM-VRN and MIMO-VRN have eight predictive modules (Fig.~\ref{fig:Architecture}) replicated and stacked for a better prediction of the missing high-frequency component. The sliding window size for LSTM-VRN is set to 7, which includes the current LR video frame together with 6 neighbouring LR frames (3 from the past and 3 from the future). The GoF size $g$ for MIMO-VRN is set to 5. For data augmentation, we randomly crop training videos to 144~$\times$~144 as HR inputs and use their bicubic-downscaled versions (of size 36~$\times$~36) as LR ground-truths. We also apply random horizontal and vertical flipping. LSTM-VRN and MIMO-VRN share the same LR and HR training objectives (Eq.~\eqref{equ:lr_loss} and Eq.~\eqref{equ:hr_loss}), with the $\lambda$ for $\mathcal{L}_{LR}$ set to 64. The training of MIMO-VRN additionally includes the center loss (Eq.~\eqref{equ:center_loss}), the hyper-parameter of which is chosen to be 16. We use Adam optimizer \cite{DBLP:journals/corr/KingmaB14}, with $\beta_1 = 0.9$, $\beta_2 = 0.5$ and a batch size of 16. The weight decay is set to $1\times10^{-12}$. We use an initial learning rate of $1\times10^{-4}$, which is decreased by half for every $30k$ iterations. Our code is available online~\footnote{\label{note1}\href{https://ding3820.github.io/MIMO-VRN/}{https://ding3820.github.io/MIMO-VRN/}}.

\input{tables/Vimeo_90K_results}

\noindent\textbf{Baselines.}
We include three categories of baselines for comparison: (1) SISO-down-SISO-up with predefined downscaling kernels (e.g.~DRN-L~\cite{guo2020closed}), (2) SISO-down-SISO-up with jointly optimized downscaling and upscaling (e.g.~CAR~\cite{sun2020learned} and IRN~\cite{xiao2020invertible}), and (3) SISO-down-MISO-up with predefined downscaling kernels (e.g.~DUF~\cite{jo2018deep}, EDVR-L~\cite{wang2019edvr}, PFNL~\cite{yi2019progressive}, TGA~\cite{isobe2020video}, and RSDN~\cite{isobe2020vide}). The first two categories perform video downscaling and upscaling on a frame-by-frame basis. The third category includes the state-of-the-art video super-resolution methods, where the predefined downscaling is done frame-by-frame and the learned upscaling is MISO-based. The predefined downscaling uses the bicubic interpolation method. It is to be noted that the methods adopting the learned downscaling perform upscaling based on their respective LR videos, which would not be the same as the bicubic-downscaled videos. The results for the methods in categories (1) and (2) are produced using the pre-trained models released by the authors. Those in category (3) are taken from the papers since these baselines share exactly the same setting as ours. We report results for a downscaling/upscaling factor of 4 only, following the common setting for video rescaling.

\noindent\textbf{Metrics.}
For quantitative comparison, we adopt the standard test protocol in the super-resolution tasks to evaluate Peak Signal-to-Noise Ratio (PSNR) and Structural Similarity Index (SSIM)~\cite{wang2004image} on the Y channel, denoted respectively by PSNR-Y and SSIM-Y.  

\begin{figure*}[t]
\begin{center}
\includegraphics[width=1.0\textwidth]{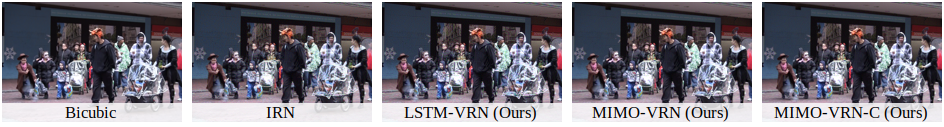}
\end{center}
\vspace{-4mm}
\caption{Sample LR video frames from Vid4. Our models show comparable visual quality to the bicubic method.}
\label{fig:lrbm}
\end{figure*}

\subsection{Comparison of Quantitative Results}
\label{sec:discussion}

Tables~\ref{table:Vid4} and \ref{table:Vimeo} report the PSNR-Y and SSIM-Y results of the reconstructed HR videos on Vid4 and Vimeo-90K-T. Table~\ref{table:LR} summarizes the results for the downscaled videos. The following observations are immediate:

(1) \textit{Optimizing jointly video downscaling and upscaling improves the HR reconstruction quality.} 
This is confirmed by the fact that LSTM-VRN achieves considerably higher PSNR-Y (32.24dB on Vid4 and 41.42dB on Vimeo-90K-T) than the baselines with video super-resolution methods for upscaling (27.33-27.92dB on Vid4 and 36.37-37.59dB on Vimeo-90K-T)~\cite{wang2019edvr, jo2018deep, yi2019progressive, isobe2020video, isobe2020vide}, which adopt the same SISO-down-MSIO-up strategy yet with a predefined downscaling kernel. We note that the image-based joint optimization schemes, e.g. IRN~\cite{xiao2020invertible} and CAR~\cite{sun2020learned}, achieve better HR reconstruction quality than the traditional video-based baselines, even without using temporal information for upscaling. The superior performance of joint optimization schemes is attributed to the fact that they can better embed HR information in LR frames for upscaling.




(2) \textit{Incorporating temporal information in the LR video improves further on the HR reconstruction quality.} The result is evidenced by the 0.95dB and 0.59dB PSNR-Y gains of LSTM-VRN over IRN~\cite{xiao2020invertible} on Vid4 and Vimeo-90K-T. Both share a similar invertible network for downscaling, but our LSTM-VRN additionally leverages information from multiple LR video frames to predict the high-frequency component of a video frame during upscaling.  



(3) \textit{MIMO-VRN achieves the best PSNR-Y/SSIM-Y results.}
It outperforms LSTM-VRN by 1.55dB and 1.84dB in PSNR-Y on Vid4 and Vimeo-90K-T, respectively, while LSTM-VRN already shows a significant improvement over the other baselines. The inclusion of the center loss (see MIMO-VRN-C) causes a modest decrease in PSNR-Y/SSIM-Y but helps to alleviate the quality fluctuation in both the resulting LR and HR videos (Sec.~\ref{sec:ablation_study}). These results highlight the benefits of incorporating temporal information into both downscaling and upscaling in an end-to-end optimized manner. 


(4) \textit{Both LSTM-VRN and MIMO-VRN produce visually-pleasing LR videos.} Table~\ref{table:LR} shows that the LR videos produced by our models have a PSNR-Y of more than 40dB when compared against the bicubic-downscaled videos. This together with the SSIM-Y results suggests that they are visually comparable to the bicubic-downscaled videos, as is also confirmed by the subjective quality comparison in Fig.~\ref{fig:lrbm} and the supplementary document.


\begin{figure*}[t]
\centering
\subfigure{
\centering
\includegraphics[scale=0.2913]{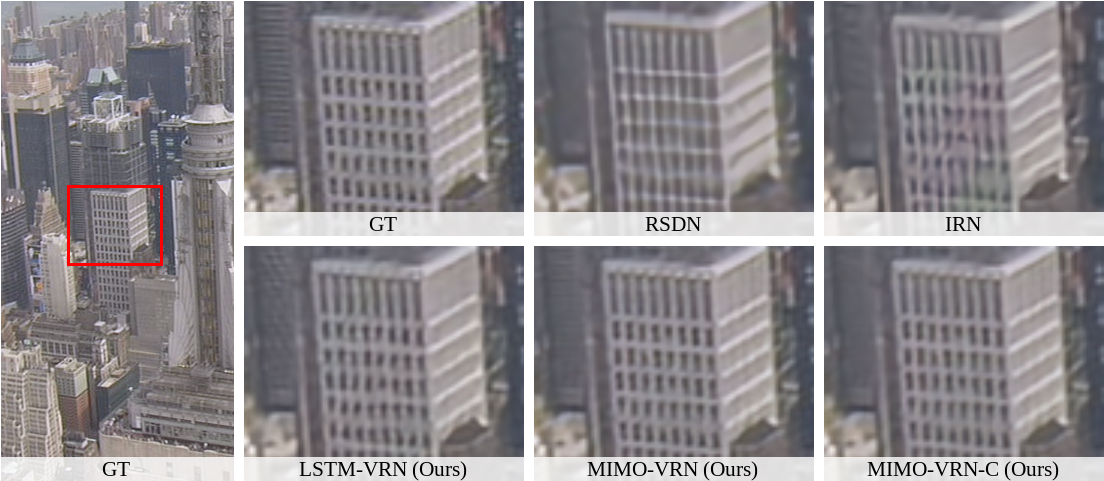}
}
\subfigure{
\centering
\includegraphics[scale=0.2913]{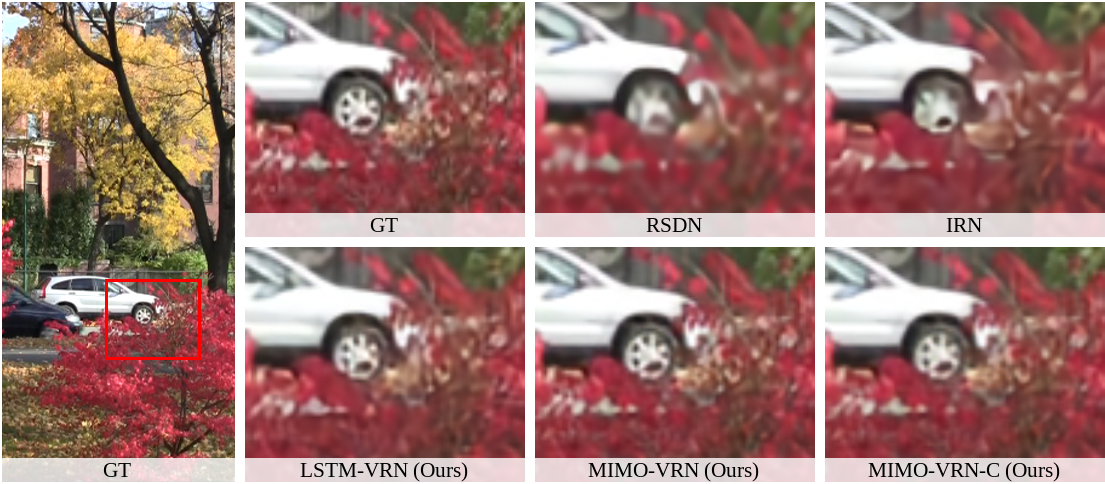}
}
\caption{Qualitative comparison on Vid4 for 4$\times$ upscaling. Zoom in for better visualization.}
\vspace{-2mm}
\label{fig:vid4}
\end{figure*}


\subsection{Comparison of Qualitative Results}
Figs.~\ref{fig:vid4} presents a qualitative comparison on Vid4. As shown, our models produce higher-quality HR video frames with much sharper edges and finer details. The other methods show blurry image quality and fail to recover image details. From Fig.~\ref{fig:lrbm}, our downscaling models produce visually comparable results to the bicubic downscaling method, which indicates the visually-pleasing property of our LR videos. The reader is referred to our project page~\textsuperscript{\ref{note1}} for more results.

\input{tables/LR}

\subsection{Ablation Experiments}
\label{sec:ablation_study}


\noindent\textbf{Temporal Propagation Methods in LSTM-VRN.}
Table~\ref{table:lstm} presents results for three temporal propagation schemes in LSTM-VRN. The first runs LSTM in forward direction without reset. The second and the third implement the proposed method with uni- or bi-directional propagation, respectively. We see that the sliding window-based reset is advantageous to the HR reconstruction quality. This may be attributed to the fact that the training videos in Vimeo-90K are rather short. When trained on Vimeo-90K, the first variant may not generalize well to unseen long videos in Vid4. As expected, with the access to both the past and future LR frames, the bi-directional propagation performs better than the uni-directional one (i.e.~Fig.~\ref{fig:LSTM} without the backward path).   

\input{tables/lstm}
\input{tables/GOP_size}
\noindent\textbf{GoF Size.} 
Table~\ref{table:gop_size} studies the effect of the GoF size on MIMO-VRN's performance. The setting GoF1 reduces to the SISO-up-SISO-down method, which is similar to IRN~\cite{xiao2020invertible} except that it introduces a prediction of the high-frequency component from the LR video frame. For a fair comparison, we re-train IRN~\cite{xiao2020invertible} on Vimeo-90K and denote the re-trained model by IRN\_Ret. Note that the pre-trained IRN~\cite{xiao2020invertible} performs better than IRN\_Ret since it is trained on a different (image-based) dataset. We see that GoF1 and IRN\_Ret show comparable performance, especially on the HR videos. This suggests that without additional temporal information, the prediction of the high-frequency component from the LR video is ineffective. However, increasing the GoF size, which involves more temporal information in downscaling and upscaling, improves the quality of the HR video significantly. GoF5 is seen to be the best setting.  




\begin{figure}[t]
\centering
\subfigure[Reconstructed HR video]{
\centering
\includegraphics[width=.226\textwidth]{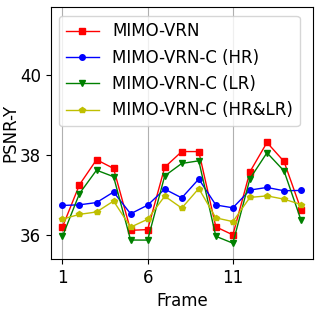}
\label{fig:hrcl-a}
}
\subfigure[Downscaled LR video]{
\centering
\includegraphics[width=.226\textwidth]{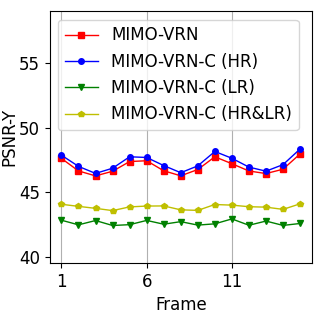}
\label{fig:lrcl-b}
}
\caption{The impact of the center loss on the quality of the HR and LR videos. The per-frame PSNR-Y is visualized as a function of frame indices. The GoF size is 5. \textbf{MIMO-VRN}: no center loss. \textbf{HR}: the center loss imposed on the HR video only. \textbf{LR}: the center loss imposed on the LR video only. \textbf{HR\&LR}: the center loss imposed on both the HR and LR videos.}
\label{fig:centerloss}
\end{figure}
\input{tables/Center_loss}

\noindent\textbf{Center Loss.} 
Fig.~\ref{fig:centerloss} visualizes the PSNR-Y of the HR and LR videos produced by MIMO-VRN as functions of time. Without the center loss (see MIMO-VRN), the PSNR-Y of both the HR and LR videos fluctuates periodically by as much as 2dB. Observe that the crest points of the HR video occur roughly at the GoF centers while the trough points are at the GoF boundaries. Table~\ref{table:centerloss} performs an ablation study of how this center loss would affect the HR and/or LR videos when it is imposed on these videos. We observe that introducing the center loss largely mitigates the quality fluctuation in the corresponding HR and/or LR video (see the MAD results in Table~\ref{table:centerloss} and Fig.~\ref{fig:centerloss}). It however degrades the HR and/or LR quality in terms of PSNR-Y, as compared to the case without the loss. We make the choice of imposing the center loss on the HR video only for two reasons. First, this leads to a minimal impact on the HR reconstruction quality. The second is that the quality fluctuation in the LR video is less problematic in terms of subjective quality because the PSNR-Y measured against the bicubic-downscaled video is way above 40dB. On closer visual inspection, these LR videos hardly show any artifacts in the temporal dimension.

\vspace{-3mm}
\subsection{Complexity-performance Trade-offs}
LSTM-VRN and MIMO-VRN present different complexity-performance trade-offs. (1) LSTM-VRN is relatively lightweight, having 9M network parameters as compared to 19M with MIMO-VRN. (2) LSTM-VRN does not require additional buffering/delay and storage for downscaling as is necessary for MIMO-VRN. (3) MIMO-VRN has better LR/HR quality while LSTM-VRN has more consistent LR/HR quality temporally. They use depends on the complexity constraints and performance requirements of the application.

%% file: tables/vid4_results.tex
\begin{table*}[t]\centering
\setlength{\tabcolsep}{4.0pt}
\caption{PSNR-Y / SSIM-Y comparison on Vid4 for $\times4$ upscaling. '$\dagger$' represents the model adopting the joint optimization for downscaling and upscaling. \textcolor{red}{Red}, \textcolor{ForestGreen}{green}, and \textcolor{blue}{blue} indicate the best, the second best, and the third best performance, respectively.}

\begin{tabular}{cccccccc}
\hline
Downscale & Upscale & Method & Calendar  & City  & Foliage  & Walk  & Average \\ \hline \hline
\multirow{9}{*}{SISO} 
& \multirow{3}{*}{SISO} 
& DRN-L~\cite{guo2020closed} & 22.47 / 0.7289 & 26.25 / 0.7011 & 24.88 / 0.6681 & 28.84 / 0.8752 & 25.61 / 0.7433 \\
&& CAR$^\dagger$~\cite{sun2020learned} & 24.48 / 0.8143 & 30.19 / 0.8444 & 26.98 / 0.7841 & 31.59 / 0.9250 & 28.28 / 0.8421 \\
&& IRN$^\dagger$~\cite{xiao2020invertible} & 26.62 / 0.8850 & 33.48 / 0.9337 & 29.71 / 0.8871 & 35.36 / 0.9696 & 31.29 / 0.9188 \\\cline{2-8}
& \multirow{6}{*}{MISO}
& DUF~\cite{jo2018deep} & 24.04 / 0.8110 & 28.27 / 0.8313 & 26.41 / 0.7709 & 30.60 / 0.9141 & 27.33 / 0.8318 \\
&& EDVR-L~\cite{wang2019edvr} & 24.05 / 0.8147 & 28.00 / 0.8122 & 26.34 / 0.7635 & 31.02 / 0.9152 & 27.35 / 0.8264 \\
&& PFNL~\cite{yi2019progressive} & 24.37 / 0.8246 & 28.09 / 0.8385 & 26.51 / 0.7768 & 30.65 / 0.9135 & 27.40 / 0.8384 \\
&& TGA~\cite{isobe2020video} & 24.47 / 0.8286 & 28.37 / 0.8419 & 26.59 / 0.7793 & 30.96 / 0.9181 & 27.59 / 0.8419 \\
&& RSDN~\cite{isobe2020vide} & 24.60 / 0.8355 & 29.20 / 0.8527 & 26.84 / 0.7931 & 31.04 / 0.9210 & 27.92 / 0.8505 \\
&& LSTM-VRN$^\dagger$ & \textcolor{blue}{27.31 / 0.9039} & \textcolor{blue}{34.36 / 0.9482} & \textcolor{blue}{31.13 / 0.9213} & \textcolor{blue}{36.18 / 0.9742} & \textcolor{blue}{32.24 / 0.9369} \\
\hline
\multirow{2}{*}{MIMO}
& \multirow{2}{*}{MIMO} &
MIMO-VRN$^\dagger$ & \textcolor{red}{29.23 / 0.9389} & \textcolor{red}{35.49 / 0.9573} & \textcolor{red}{33.25 / 0.9535} & \textcolor{red}{37.17 / 0.9812} & \textcolor{red}{33.79 / 0.9577} \\
&& MIMO-VRN-C$^\dagger$ & \textcolor{ForestGreen}{28.83 / 0.9322} & \textcolor{ForestGreen}{35.13 / 0.9544} & \textcolor{ForestGreen}{32.72 / 0.9476} & \textcolor{ForestGreen}{36.93 / 0.9808} & \textcolor{ForestGreen}{33.40 / 0.9537} \\ \hline

\end{tabular}
\label{table:Vid4}
\end{table*}

%% file: tables/Vimeo_90K_results.tex
\begin{table}[t]\centering
\setlength{\tabcolsep}{4.5pt}
\caption{PSNR-Y / SSIM-Y comparison on Vimeo-90K-T for $\times4$ upscaling. '$\dagger$' represents the model adopting the joint optimization for downscaling and upscaling. \textcolor{red}{Red}, \textcolor{ForestGreen}{green}, and \textcolor{blue}{blue} indicate the best, the second best, and the third best performance, respectively.}
\begin{tabular}{cccc}
\hline
Downscale & Upscale & Method & Average \\ \hline \hline
\multirow{9}{*}{SISO} 
& \multirow{3}{*}{SISO} 
& DRN-L~\cite{guo2020closed} & 35.63 / 0.9262 \\
&& CAR$^\dagger$~\cite{sun2020learned} & 37.69 / 0.9493 \\
&& IRN$^\dagger$~\cite{xiao2020invertible} &  40.83 / 0.9734 \\\cline{2-4}
& \multirow{5}{*}{MISO}
& DUF~\cite{jo2018deep} & 36.37 / 0.9387\\
&& EDVR-L~\cite{wang2019edvr} & 37.63 / 0.9487\\
&& TGA~\cite{isobe2020video} & 37.59 / 0.9516\\
&& RSDN~\cite{isobe2020vide} & 37.23 / 0.9471\\
&& LSTM-VRN$^\dagger$ & \textcolor{blue}{41.42} / \textcolor{blue}{0.9764} \\
\hline
\multirow{2}{*}{MIMO}
& \multirow{2}{*}{MIMO} 
& MIMO-VRN$^\dagger$ & \textcolor{red}{43.26} / \textcolor{red}{0.9846}\\ 
&& MIMO-VRN-C$^\dagger$ & \textcolor{ForestGreen}{42.53} / \textcolor{ForestGreen}{0.9820}\\ 
\hline
\end{tabular}
\vspace{-3mm}
\label{table:Vimeo}
\end{table}

%% file: tables/LR.tex
\begin{table}[t]\centering
\setlength{\tabcolsep}{5.8pt}
\caption{PSNR-Y and SSIM-Y results measured between the $\times4$ downscaled LR videos and the bicubic-downscaled videos.}
\begin{tabular}{c|c|c}
\hline
Method & Vid4 & Vimeo-90K-T \\ \hline
IRN~\cite{xiao2020invertible} & 40.77 / 0.9908 & 46.24 / 0.9956 \\
LSTM-VRN & 42.36 / 0.9940 & 47.14 / 0.9968\\
MIMO-VRN & 45.05 / 0.9965 & 49.11 / 0.9975\\
MIMO-VRN-C & 45.51 / 0.9969 & 49.34 / 0.9976\\
\hline
\end{tabular}
\vspace{-2mm}
\label{table:LR}
\end{table}

%% file: tables/lstm.tex
\begin{table}[t]\centering
\setlength{\tabcolsep}{5.8pt}
\caption{Ablation study of the propagation methods for LSTM-VRN. Results are reported on Vid4.}
\begin{tabular}{c c|c}
\hline
Sliding Window & Bi-directional & PSNR-Y\\ \hline
& & 31.16 \\
$\surd$& & 31.53 \\
$\surd$ &$\surd$& 32.24\\ \hline

\end{tabular}
\label{table:lstm}
\end{table}

%% file: tables/GOP_size.tex
\begin{table}[t]\centering
\setlength{\tabcolsep}{5.8pt}
\caption{PSNR-Y of different GoF sizes on Vid4. IRN\_Ret is the re-trained IRN with Vimeo-90K, as compared to IRN, the pre-trained model from~\cite{xiao2020invertible}.}
\begin{tabular}{c|c|c}
\hline
Method & HR & LR \\ \hline
IRN~\cite{xiao2020invertible} & 31.29 & 41.13 \\
IRN\_Ret & 30.72 & 45.06 \\
GoF1 & 30.69  & 44.38 \\
GoF3 & 33.61  & 43.85 \\
GoF5 & 33.79  & 45.05 \\
GoF7 & 33.45  & 45.13 \\
\hline
\end{tabular}
\label{table:gop_size}
\end{table}

%% file: tables/Center_loss.tex
\begin{table}[t]
\centering
\setlength{\tabcolsep}{5.8pt}
\caption{PSNR-Y of MIMO-VRN with and without the center loss on Vid4. The mean absolute deviation (MAD) indicates the average absolute deviation of the per-frame PSNR-Y from the GoF mean.} 


\begin{tabular}{cc|cc|cc}
\hline
\multicolumn{2}{c|}{Center loss} & \multicolumn{2}{c|}{PSNR-Y} & \multicolumn{2}{c}{MAD} \\
\cline{1-6}
HR & LR & HR & LR & HR & LR \\

\hline
 & &  33.79 & 45.05 & 0.88 & 0.55\\
 $\surd$& & 33.40 & 45.54 & 0.28 & 0.63\\
 & $\surd$& 33.55 & 42.42 & 0.86 &0.38\\
 $\surd$& $\surd$&  33.13 & 43.32 & 0.31& 0.29\\
\hline

\end{tabular}
\label{table:centerloss}
\end{table}

%% file: conclusion.tex
\section{Conclusion}
This work presents two joint optimization approaches to video rescaling. Both incorporate an invertible network with coupling layer architectures to model explicitly the high-frequency component inherent in the HR video. While our LSTM-VRN shows that the temporal information in the LR video can be utilized to good advantage for better upscaling, our MIMO-VRN demonstrates that the GoF-based rescaling is able to make full use of temporal information to benefit both upscaling and downscaling. Our models demonstrate superior quantitative and qualitative performance to the image-based invertible model. They outperform, by a significant margin, the video rescaling framework without joint optimization.  

%% file: appendix.tex

This supplementary document provides additional results to validate the proposed methods. These include (1) quantitative results based on the Video Multi-Method Assessment Fusion (VMAF) metric~\cite{7986143}, a video quality metric that is shown to correlate highly with human perception; (2) an examination of temporal consistency on the downscaled and upscaled videos following the method in RSDN~\cite{isobe2020vide}; (3) an ablation study of the predictive module; and (4) more qualitative results of the downscaled and upscaled videos.   



\section{Quantitative results based on VMAF}
\label{sec:VMAF}
This section presents quantitative results for upscaled and downscaled videos in Vid4 based on the Video Multi-Method Assessment Fusion (VMAF)~\cite{7986143} metric. VMAF is an objective video quality metric that is shown to correlate highly with human perception. It accepts as inputs a reference video and a distorted video. Its output is a score between 0 and 100. The higher the VMAF score, the more closely the two input videos match each other. Compared in Table~\ref{table:vmaf} are the VMAF scores of the reconstructed high-resolution (HR) videos produced by a few baselines (whose code is available) and our schemes. The reference videos are the original HR videos. Likewise, Table~\ref{table:vmaf_lr} shows the results for the downscaled videos, where the reference videos are the bicubic-downscaled videos.   

From Table~\ref{table:vmaf}, we observe that our LSTM-VRN, MIMO-VRN and MIMO-VRN-C outperform IRN~\cite{xiao2020invertible} (image-based joint optimization scheme) consistently. As compared with EDVR-L~\cite{wang2019edvr}, a traditional video super-resolution approach, the proposed methods show significantly improved VMAF scores. These observations are in line with the PSNR/SSIM results reported in Table 1 of the main paper. A similar observation can be made regarding the results of the downscaled LR videos in Table~\ref{table:vmaf_lr}, which agrees with the PSNR/SSIM results in Table 3 of the main paper. Remarkably, all three proposed methods have VMAF scores very close to 100, suggesting that their LR videos are visually similar to bicubic-downscaled videos. 


\input{tables/vmaf}
\input{tables/vmaf_lr}

\input{tables/zeros_v2}
\section{Temporal consistency}
\label{sec:tc}
\begin{figure*}[t]
\centering
\subfigure{
\centering
\includegraphics[scale=0.45]{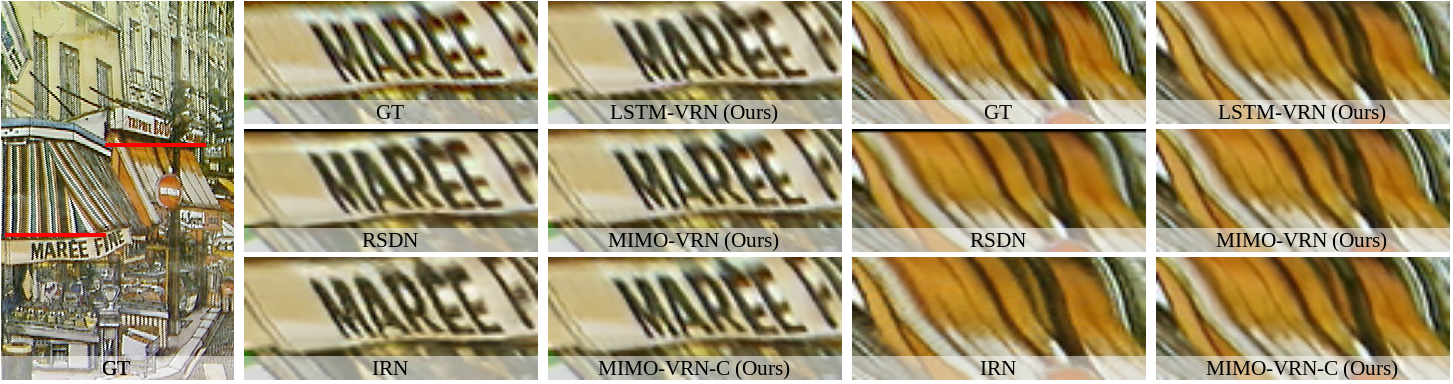}
}
\subfigure{
\centering
\includegraphics[scale=0.45]{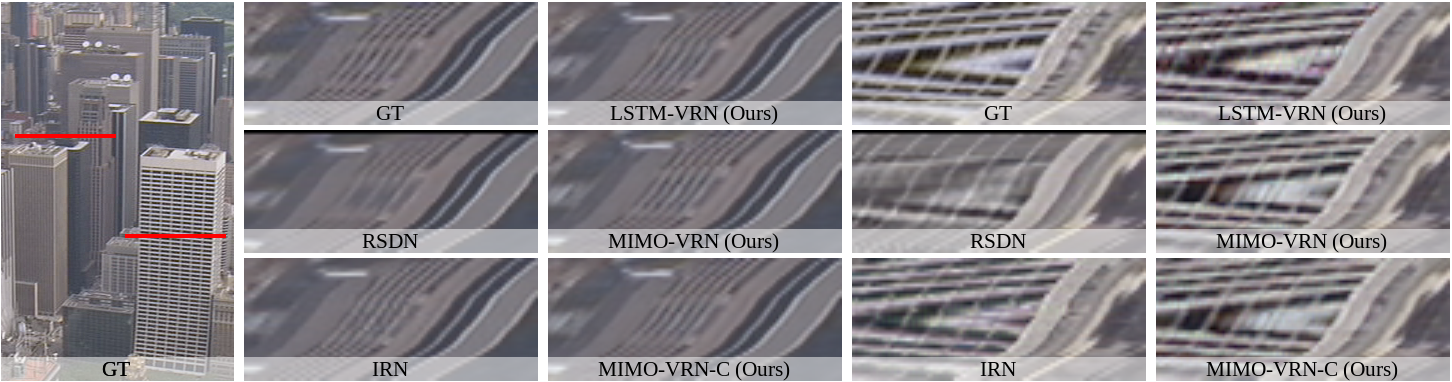}
}
\subfigure{
\centering
\includegraphics[scale=0.45]{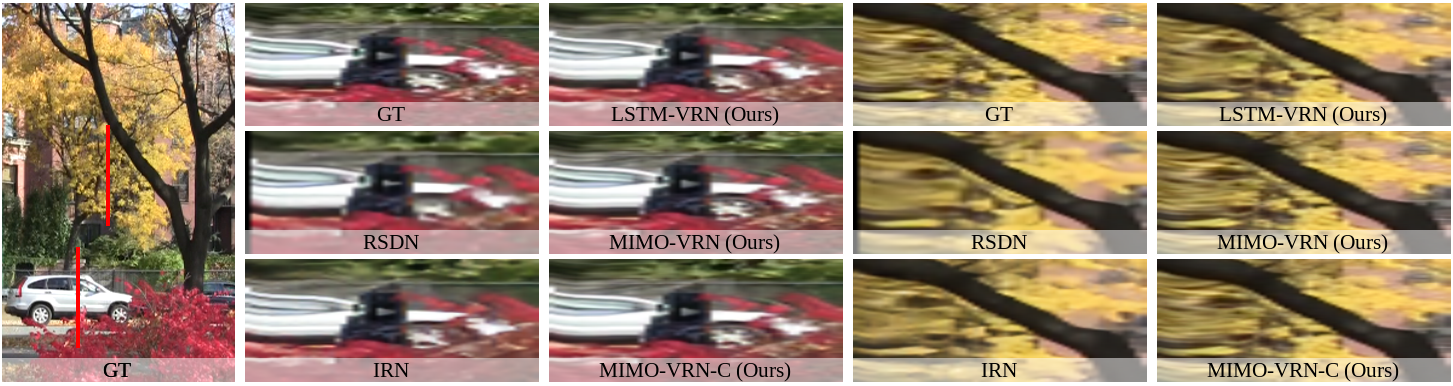}
}
\subfigure{
\centering
\includegraphics[scale=0.45]{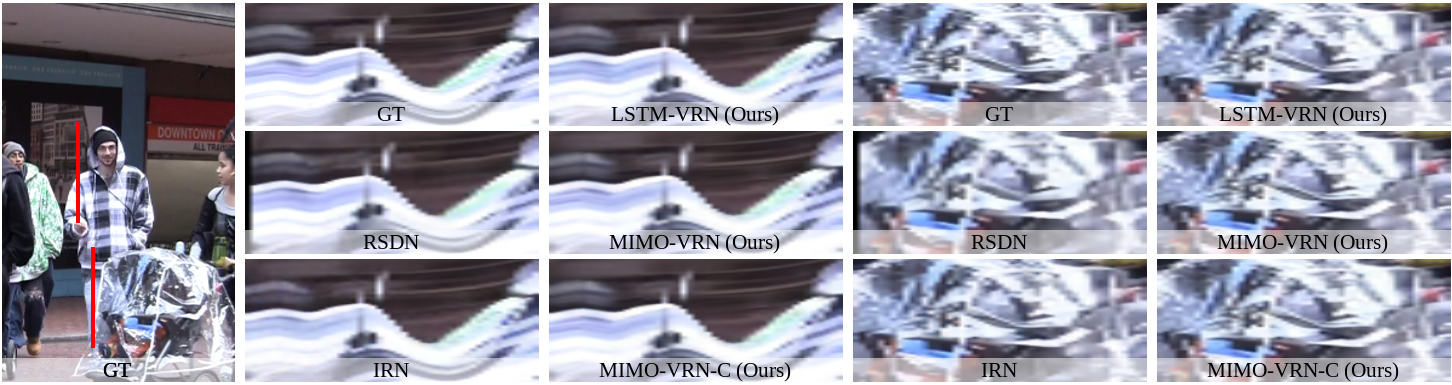}
}
\caption{Visualization of temporal consistency across the reconstructed HR video frames. The images shown are formed by vertically (or horizontally) stitching rows (or columns) of pixels extracted separately from consecutive video frames at co-located positions (indicated by the red lines).}
\label{fig:tpsr}
\end{figure*}

In this section, the temporal consistency of the downscaled and upscaled videos is examined. We follow RSDN~\cite{isobe2020vide} to extract a row or a column of pixels at the co-located positions in consecutive video frames. We then stitch vertically (or horizontally) these extracted rows (or columns) of pixels to form an image, in order to visualize their variations in the temporal dimension. From Fig.~\ref{fig:tpsr}, we are not aware of any noticeable inconsistency across reconstructed HR video frames. Moreover, the resulting images of our methods resemble closely the ground-truths.

\begin{figure*}[t]
\centering
\subfigure{
\centering
\includegraphics[scale=0.5]{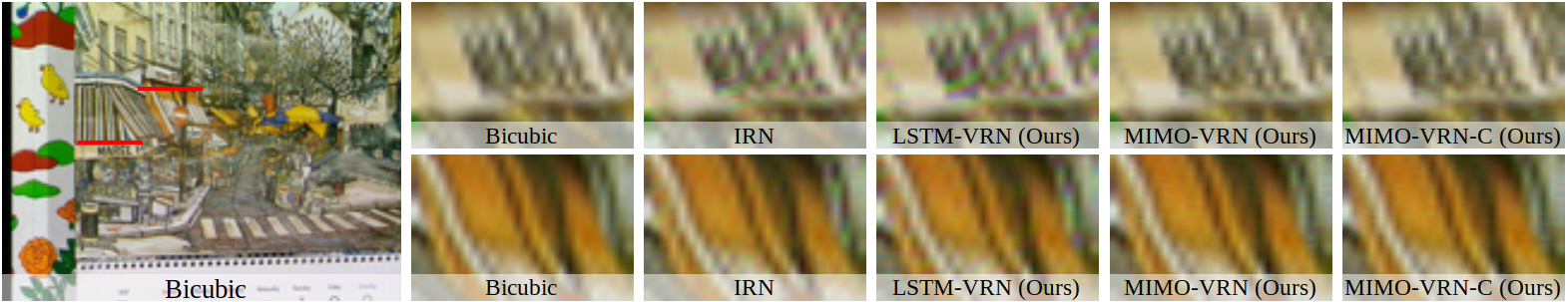}
}
\subfigure{
\centering
\includegraphics[scale=0.5]{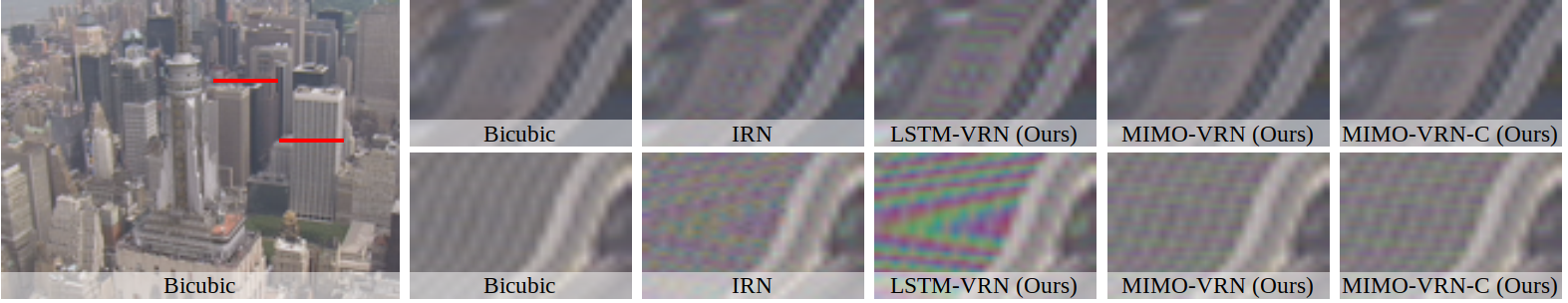}
}
\subfigure{
\centering
\includegraphics[scale=0.5]{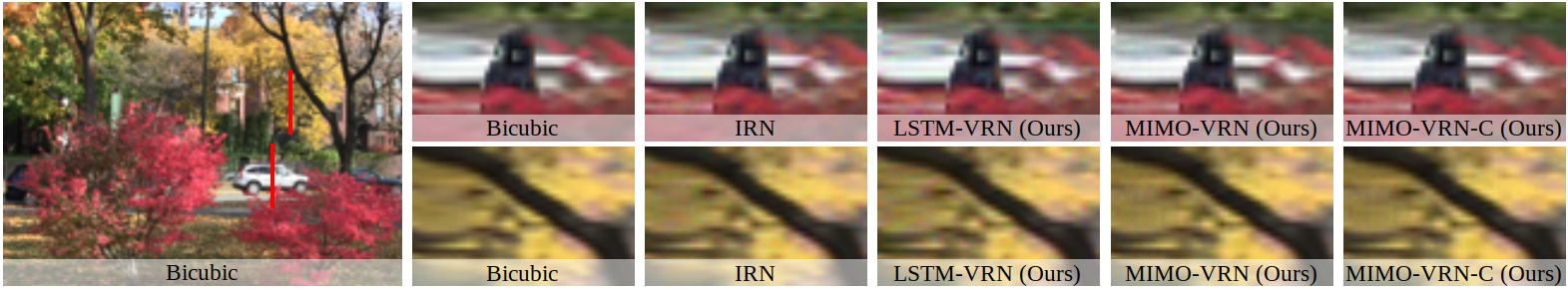}
}
\subfigure{
\centering
\includegraphics[scale=0.5]{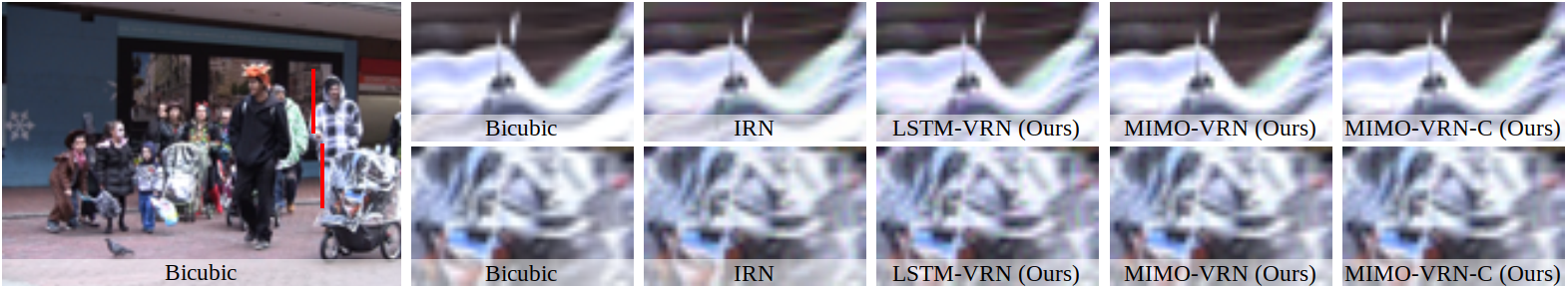}
}
\caption{Visualization of temporal consistency across the downscaled LR video frames. The images shown are formed by vertically (or horizontally) stitching rows (or columns) of pixels extracted separately from consecutive video frames at co-located positions (indicated by the red lines).}
\label{fig:tplr}
\end{figure*}

In Fig.~\ref{fig:tplr}, IRN~\cite{xiao2020invertible} and LSTM-VRN result in slight temporal inconsistency in some areas of the low-resolution (LR) videos, particularly the Calendar and City sequences. This is evidenced by the aliasing artifact especially appeared on the alphabet of Calendar sequence and the building of City sequence. Nevertheless, such inconsistency does not appear in the LR videos produced by MIMO-VRN and MIMO-VRN-C.

\section{The effectiveness of the predictive module}
\label{sec:z}
Table~\ref{table:zero} provides quantitative results to justify the effectiveness of the proposed predictive module. Recall that the predictive module forms a prediction $\hat{z}$ of the missing high-frequency component $z$ from the LR video frames $\hat{y}$. This new feature distinguishes our schemes from IRN~\cite{xiao2020invertible}, the image-based joint optimization scheme, which uses a Gaussian noise for $\hat{z}$. The upper section of Table~\ref{table:zero} compares the results of LSTM-VRN with IRN\_Ret (the re-trained IRN that uses the same training dataset as LSTM-VRN), validating that the predictive module is effective for reconstructing better HR videos. The lower section of Table~\ref{table:zero} conducts the same analysis for MIMO-VRN-C, where we replace $\hat{z}$ produced by the predictive module with a fixed zero tensor, a scheme termed MIMO-VRN-C-Zero. MIMO-VRN-C-Zero is trained in the same way as MIMO-VRN-C. We see that the predictive module is still effective, even though the gain is less significant than the case in LSTM-VRN.   



\section{More qualitative results}
\label{sec:qr}
\begin{figure*}[t]
\centering
\subfigure{
\centering
\includegraphics[scale=0.29005]{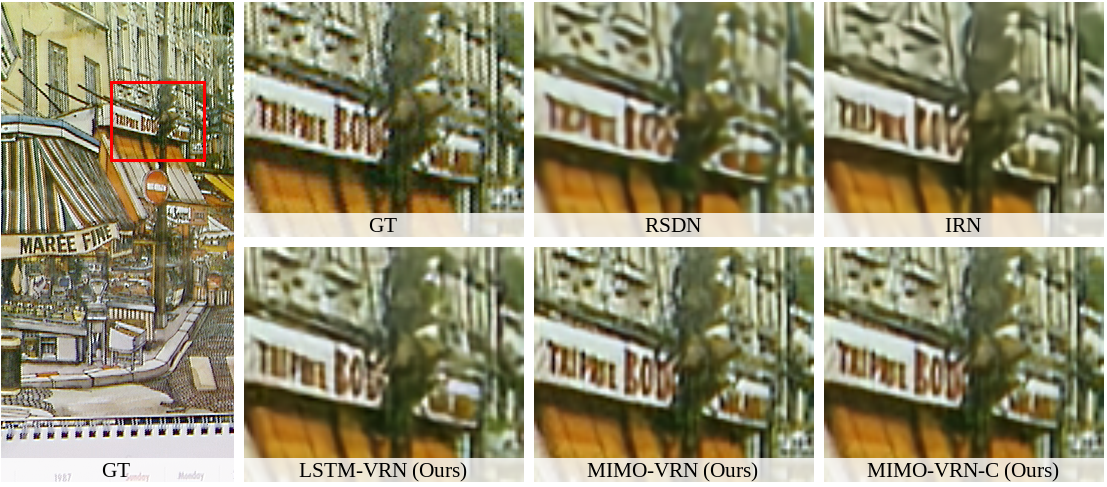}
}
\subfigure{
\centering
\includegraphics[scale=0.29005]{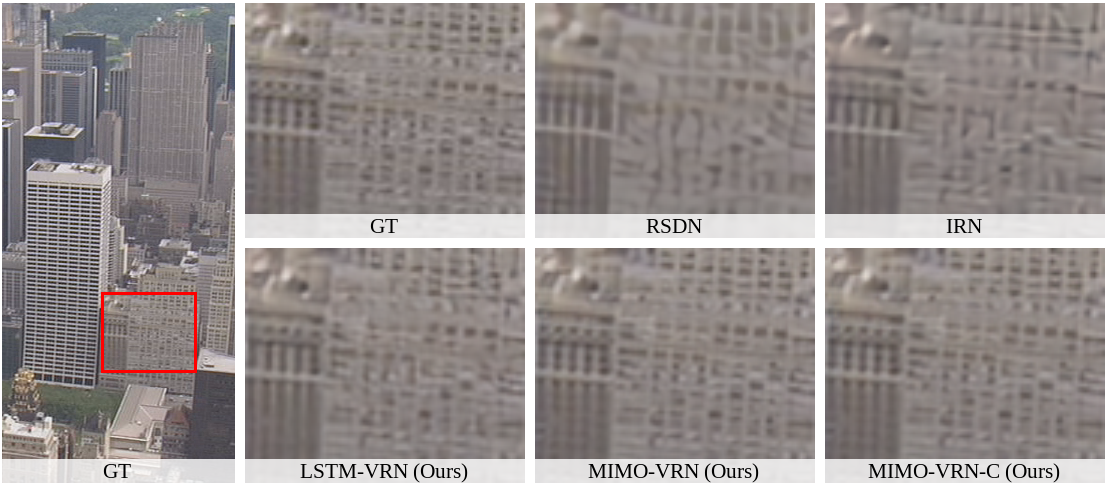}
}
\subfigure{
\centering
\includegraphics[scale=0.29005]{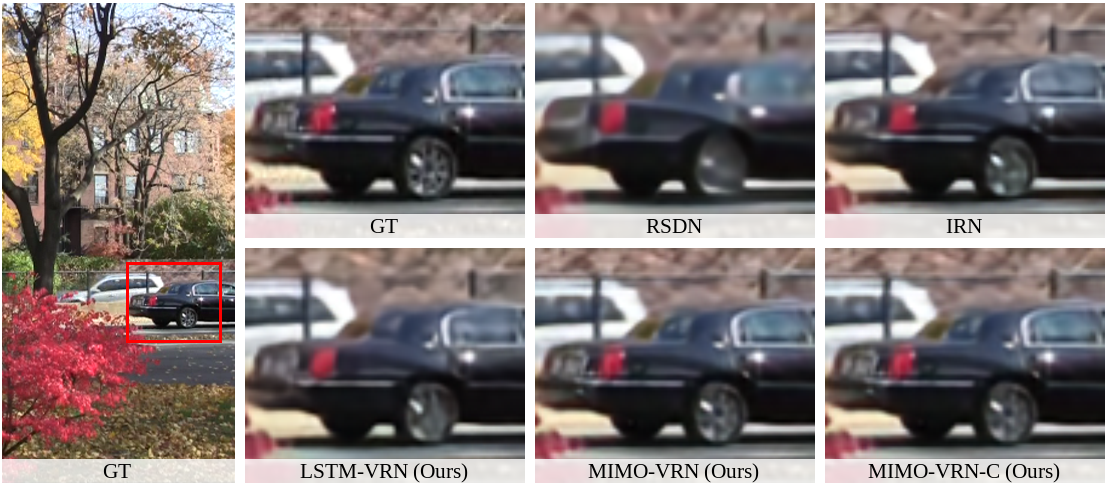}
}
\subfigure{
\centering
\includegraphics[scale=0.29005]{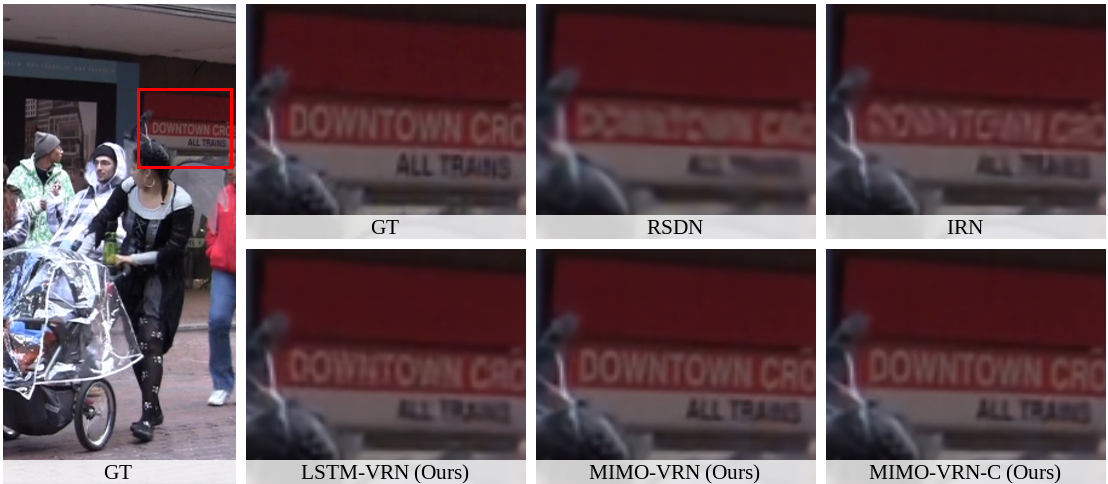}
}
\caption{Qualitative comparison on Vid4 for $\times4$ upscaling. Zoom in for better visualization. }
\label{fig:srvid}
\end{figure*}

\begin{figure*}[t]
\begin{center}
\includegraphics[width=1.0\textwidth]{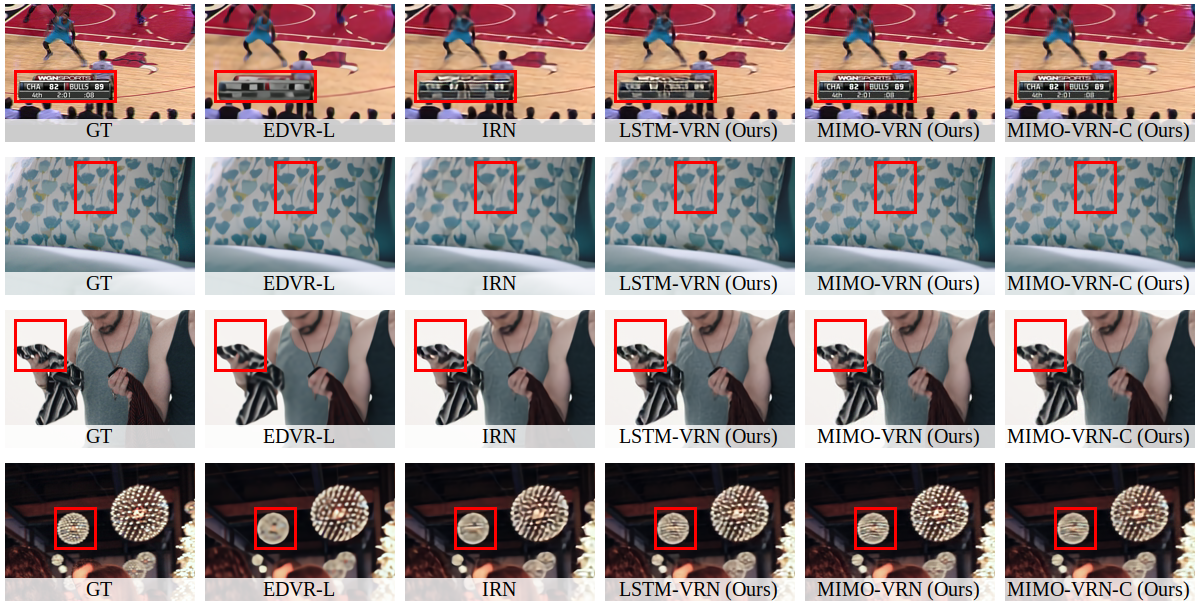}
\end{center}
\caption{Qualitative comparison on Vimeo-90K-T for $\times4$ upscaling. Zoom in for better visualization.}
\label{fig:srvim}
\end{figure*}

Figs.~\ref{fig:srvid} and \ref{fig:srvim} provide more qualitative results, comparing the reconstructed HR videos of different models. They again suggest that our models can recover fine details and sharp edges.

\begin{figure*}[t]
\centering
\subfigure{
\centering
\includegraphics[scale=0.29]{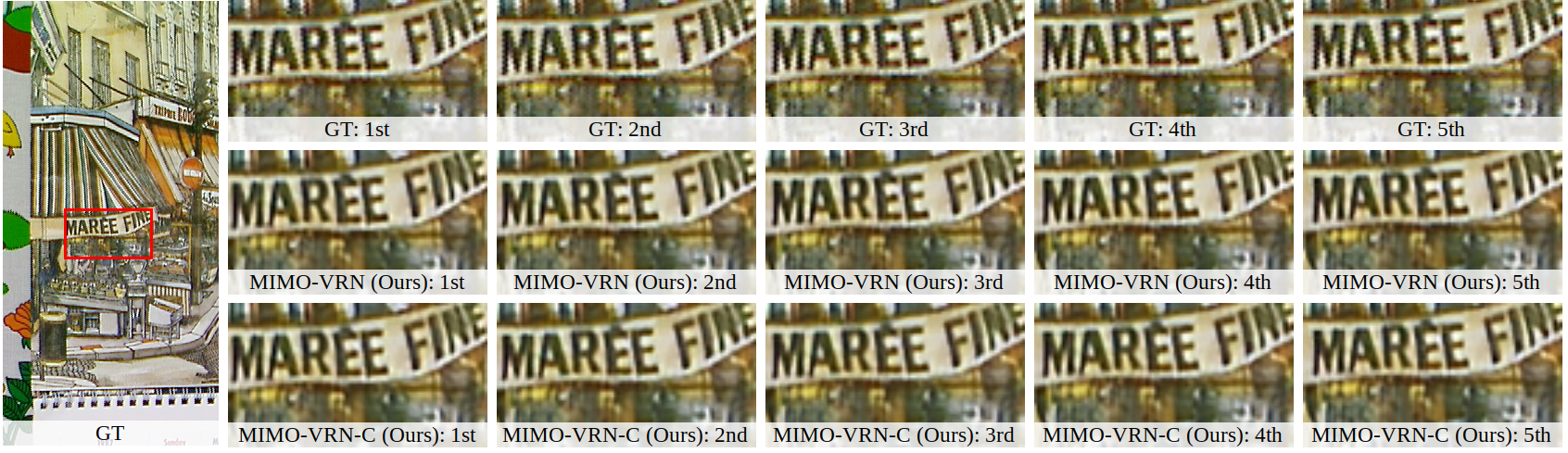}
}
\subfigure{
\centering
\includegraphics[scale=0.29]{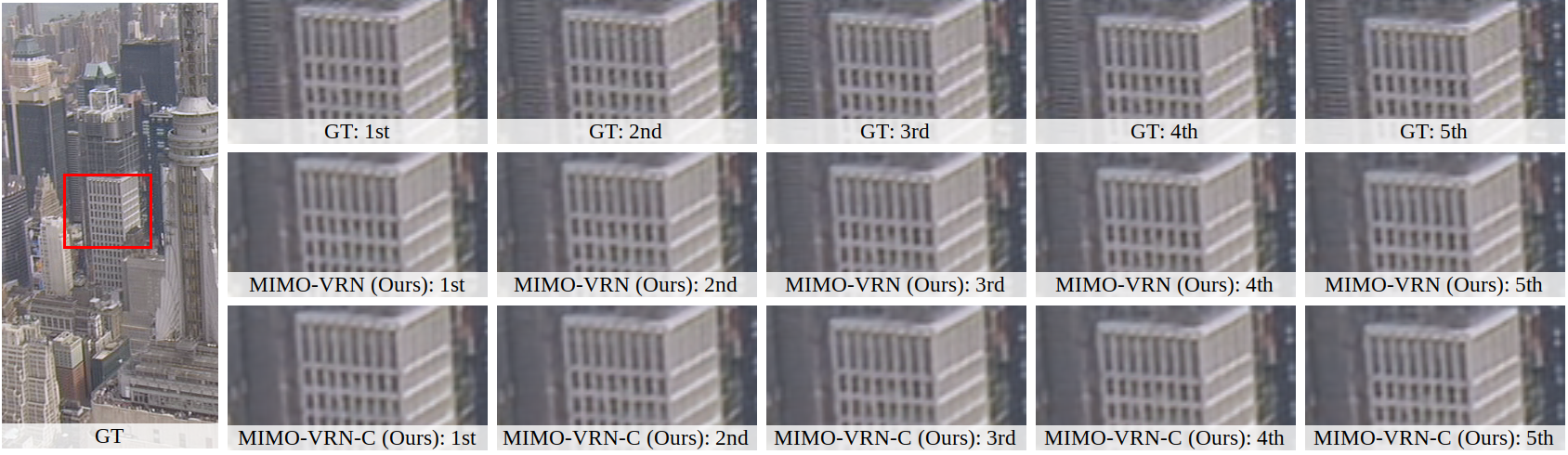}
}
\subfigure{
\centering
\includegraphics[scale=0.29]{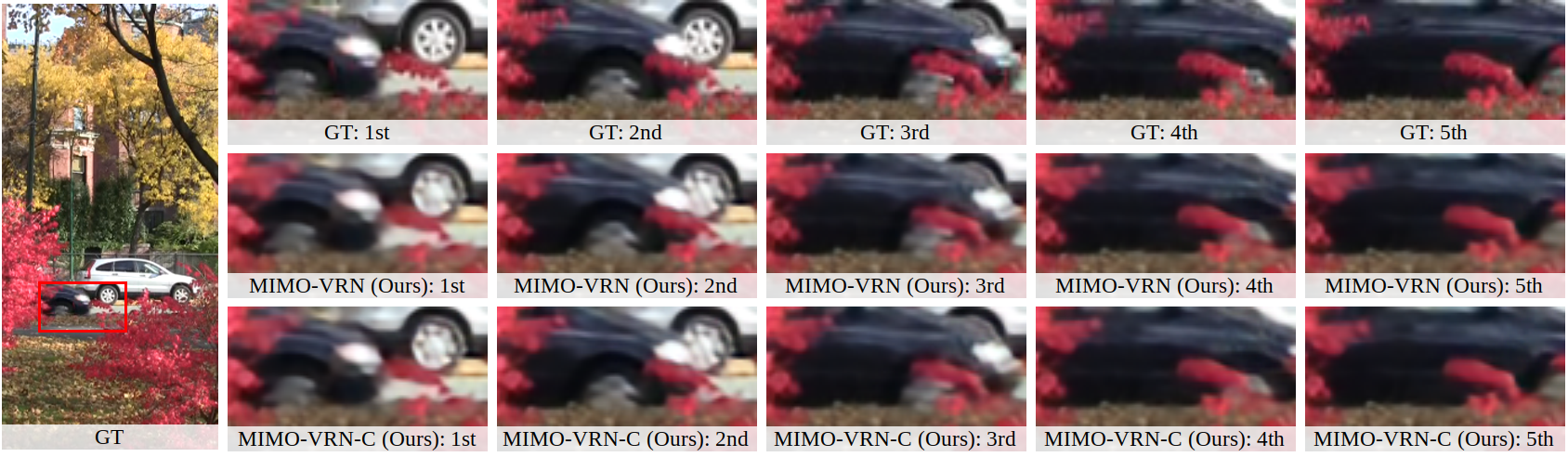}
}
\subfigure{
\centering
\includegraphics[scale=0.29]{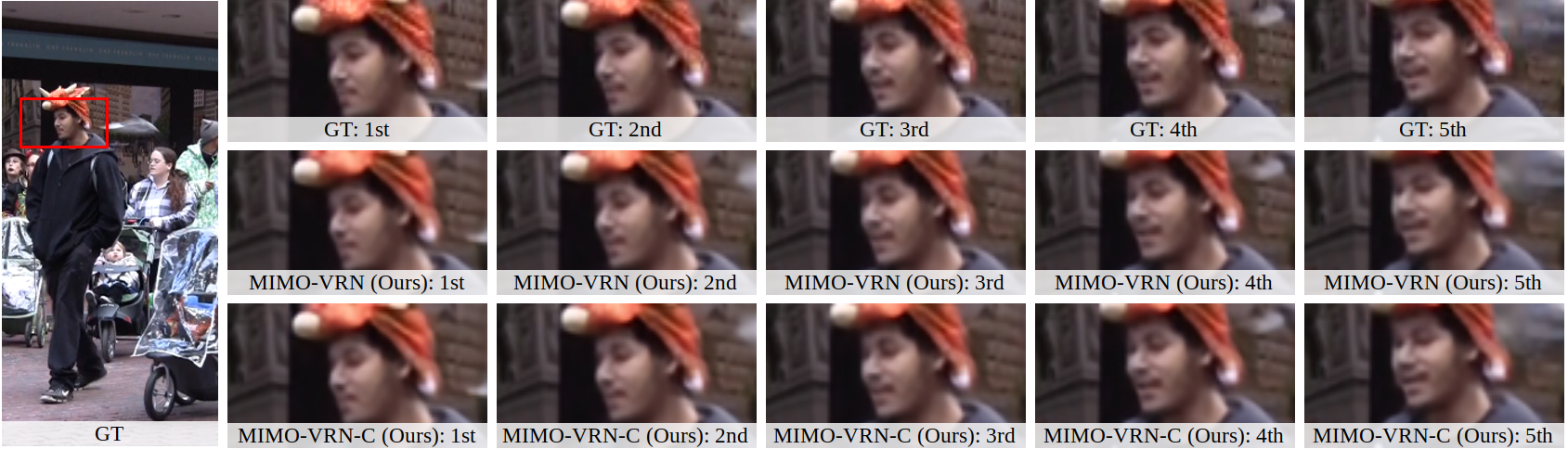}
}
\caption{Qualitative frame-by-frame comparison between MIMO-VRN and MIMO-VRN-C, with a GoF size of 5. The images shown are $\times4$ upscaled HR video frames of Vid4.}
\label{fig:tpgof5}
\end{figure*}

\begin{figure*}[t]
\begin{center}
\includegraphics[width=1.0\textwidth]{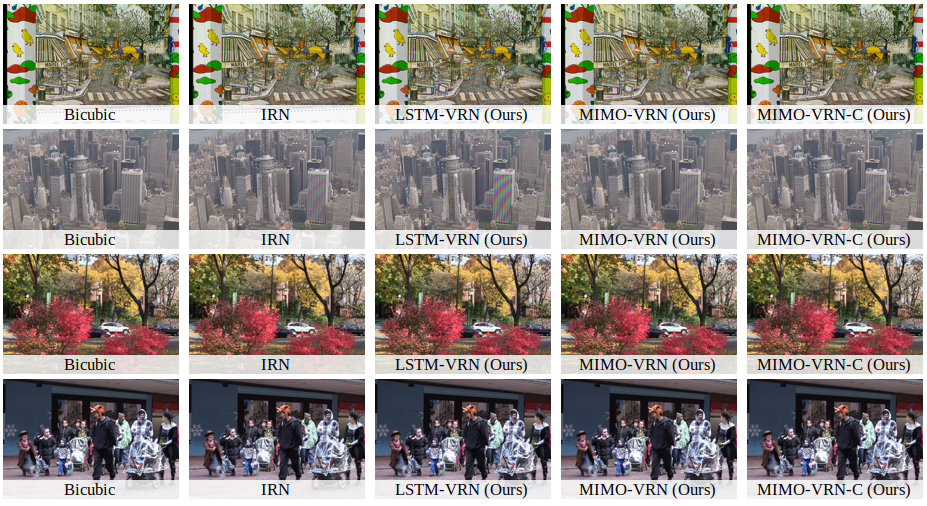}
\end{center}
\caption{Sample LR video frames from Vid4. Our models show comparable visual quality to the bicubic method.}
\label{fig:lr}
\end{figure*}

Fig.~\ref{fig:tpgof5} presents a frame-by-frame qualitative comparison between MIMO-VRN and MIMO-VRN-C, with a GoF size of 5. MIMO-VRN-C comes with an additional center loss to ensure temporal consistency. It is seen that both MIMO-VRN and MIMO-VRN-C can successfully reconstruct image details. Comparing MIMO-VRN with MIMO-VRN-C, we are not aware of any significant quality variation in the temporal dimension, even though Fig.~7 of the main paper suggests that the HR quality of MIMO-VRN may fluctuate more significantly than MIMO-VRN-C. Fig.~\ref{fig:lr} displays more LR video frames from Vid4, showing that our models offer comparable visual quality to the bicubic-downscaled videos. 



%

%

%

%

%

%

%% file: tables/vmaf.tex
\begin{table}[t]\centering
\setlength{\tabcolsep}{2.0pt}
\caption{Comparison of VMAF scores for $\times 4$ upscaled HR videos on Vid4. The reference videos are the original HR videos. The higher the VMAP scores, the better the HR reconstruction quality. \textcolor{red}{Red} and \textcolor{ForestGreen}{green} indicate the best and the second best performance, respectively.}
\begin{tabular}{c|c|c|c|c|c}
\hline
Method & Calendar  & City  & Foliage  & Walk  & Average \\ \hline
EDVR-L~\cite{wang2019edvr} & 64.11 & 47.22 & 77.12 & 89.86 & 69.58 \\
IRN~\cite{xiao2020invertible} & 77.85 & 80.80 & 89.83 & 97.01 & 86.37 \\
LSTM-VRN & 78.03 & 85.16 & 91.14 & 97.81 & 88.04 \\
MIMO-VRN & \textcolor{red}{81.99} & \textcolor{red}{87.35} & \textcolor{red}{95.31} & \textcolor{red}{98.57} & \textcolor{red}{90.81} \\
MIMO-VRN-C & \textcolor{ForestGreen}{80.44} & \textcolor{ForestGreen}{85.47} & \textcolor{ForestGreen}{94.24} & \textcolor{ForestGreen}{98.42} & \textcolor{ForestGreen}{89.64} \\
\hline
\end{tabular}
\label{table:vmaf}
\end{table}

%% file: tables/vmaf_lr.tex
\begin{table}[t]\centering
\setlength{\tabcolsep}{2.0pt}
\caption{Comparison of VMAF scores for $\times 4$ downscaled LR videos on Vid4. The reference videos are generated by the bicublic downscaling method. The higher the VMAP scores, the more closely the LR videos resemble the bicubic-downscaled ones. \textcolor{red}{Red} and \textcolor{ForestGreen}{green} indicate the best and the second best performance, respectively.}
\begin{tabular}{c|c|c|c|c|c}
\hline
Method & Calendar  & City  & Foliage  & Walk  & Average \\ \hline
IRN~\cite{xiao2020invertible} & 94.18 & 94.92 & 95.38 & 97.82 & 95.55 \\
LSTM-VRN & 95.86 & 95.31 & 96.52 & 99.47 & 96.79 \\
MIMO-VRN & \textcolor{ForestGreen}{97.37} & \textcolor{ForestGreen}{96.84} & \textcolor{ForestGreen}{97.88} & \textcolor{ForestGreen}{99.83} & \textcolor{ForestGreen}{97.98} \\
MIMO-VRN-C & \textcolor{red}{97.94} & \textcolor{red}{97.17} & \textcolor{red}{98.34} & \textcolor{red}{99.87} & \textcolor{red}{98.33} \\
\hline
\end{tabular}
\label{table:vmaf_lr}
\end{table}

%% file: tables/zeros_v2.tex
\begin{table*}[t]
\centering
\setlength{\tabcolsep}{6.5pt}
\caption{Ablation study of the predictive module. IRN\_Ret is trained using the same dataset as LSTM-VRN. MIMO-VRN-C-Zero is an implementation of MIMO-VRN-C without the predictive module. All the presented results are evaluated on Vid4 and they show that the proposed predictive module can improve the quality of reconstructed HR videos.}
\begin{tabular}{c|c|c}
\hline
Method & Predictive Module & HR (PSNR-Y / SSIM-Y / VMAF)\\ \hline
IRN\_Ret &  & 30.72 / 0.9087 / 86.37\\
LSTM-VRN & $\surd$ & 32.24 / 0.9369 / 88.04\\ \hline
MIMO-VRN-C-Zero &  & 33.04 / 0.9575 / 89.00 \\
MIMO-VRN-C & $\surd$ & 33.40 / 0.9609 / 89.64\\
\hline
\end{tabular}
\label{table:zero}
\end{table*}

%% file: main.bbl
\begin{thebibliography}{10}\itemsep=-1pt

\bibitem{caballero2017real}
Jose Caballero, Christian Ledig, Andrew Aitken, Alejandro Acosta, Johannes
  Totz, Zehan Wang, and Wenzhe Shi.
\newblock Real-time video super-resolution with spatio-temporal networks and
  motion compensation.
\newblock In {\em Proceedings of the IEEE Conference on Computer Vision and
  Pattern Recognition (CVPR)}, 2017.

\bibitem{chen2020hrnet}
Yuzhao Chen, Xi Xiao, Tao Dai, and Shu-Tao Xia.
\newblock Hrnet: Hamiltonian rescaling network for image downscaling.
\newblock In {\em IEEE International Conference on Image Processing (ICIP)},
  2020.

\bibitem{dai2019second}
Tao Dai, Jianrui Cai, Yongbing Zhang, Shu-Tao Xia, and Lei Zhang.
\newblock Second-order attention network for single image super-resolution.
\newblock In {\em Proceedings of the IEEE Conference on Computer Vision and
  Pattern Recognition (CVPR)}, 2019.

\bibitem{DBLP:journals/corr/DinhKB14}
Laurent Dinh, David Krueger, and Yoshua Bengio.
\newblock Nice: Non-linear independent components estimation.
\newblock In {\em Proceedings of International Conference on Learning
  Representations (ICLR)}, 2015.

\bibitem{DBLP:conf/iclr/DinhSB17}
Laurent Dinh, Jascha Sohl-Dickstein, and Samy Bengio.
\newblock Density estimation using real nvp.
\newblock In {\em Proceedings of International Conference on Learning
  Representations (ICLR)}, 2017.

\bibitem{guo2020closed}
Yong Guo, Jian Chen, Jingdong Wang, Qi Chen, Jiezhang Cao, Zeshuai Deng, Yanwu
  Xu, and Mingkui Tan.
\newblock Closed-loop matters: Dual regression networks for single image
  super-resolution.
\newblock In {\em Proceedings of the IEEE Conference on Computer Vision and
  Pattern Recognition (CVPR)}, 2020.

\bibitem{haris2019recurrent}
Muhammad Haris, Gregory Shakhnarovich, and Norimichi Ukita.
\newblock Recurrent back-projection network for video super-resolution.
\newblock In {\em Proceedings of the IEEE Conference on Computer Vision and
  Pattern Recognition (CVPR)}, 2019.

\bibitem{isobe2020vide}
Takashi Isobe, Xu Jia, Shuhang Gu, Songjiang Li, Shengjin Wang, and Qi Tian.
\newblock Video super-resolution with recurrent structure-detail network.
\newblock In {\em Proceedings of European Conference on Computer Vision
  (ECCV)}, 2020.

\bibitem{isobe2020video}
Takashi Isobe, Songjiang Li, Xu Jia, Shanxin Yuan, Gregory Slabaugh, Chunjing
  Xu, Ya-Li Li, Shengjin Wang, and Qi Tian.
\newblock Video super-resolution with temporal group attention.
\newblock In {\em Proceedings of the IEEE Conference on Computer Vision and
  Pattern Recognition (CVPR)}, 2020.

\bibitem{jo2018deep}
Younghyun Jo, Seoung Wug~Oh, Jaeyeon Kang, and Seon Joo~Kim.
\newblock Deep video super-resolution network using dynamic upsampling filters
  without explicit motion compensation.
\newblock In {\em Proceedings of the IEEE Conference on Computer Vision and
  Pattern Recognition (CVPR)}, 2018.

\bibitem{kim2018task}
Heewon Kim, Myungsub Choi, Bee Lim, and Kyoung Mu~Lee.
\newblock Task-aware image downscaling.
\newblock In {\em Proceedings of European Conference on Computer Vision
  (ECCV)}, 2018.

\bibitem{DBLP:journals/corr/KingmaB14}
Diederik~P Kingma and Jimmy Ba.
\newblock Adam: A method for stochastic optimization.
\newblock In {\em Proceedings of International Conference on Learning
  Representations (ICLR)}, 2015.

\bibitem{kingma2018glow}
Durk~P Kingma and Prafulla Dhariwal.
\newblock Glow: Generative flow with invertible 1x1 convolutions.
\newblock In {\em Advances in Neural Information Processing Systems (NIPS)},
  2018.

\bibitem{8100101}
Wei-Sheng Lai, Jia-Bin Huang, Narendra Ahuja, and Ming-Hsuan Yang.
\newblock Deep laplacian pyramid networks for fast and accurate
  super-resolution.
\newblock In {\em Proceedings of the IEEE Conference on Computer Vision and
  Pattern Recognition (CVPR)}, 2017.

\bibitem{li2020mucan}
Wenbo Li, Xin Tao, Taian Guo, Lu Qi, Jiangbo Lu, and Jiaya Jia.
\newblock Mucan: Multi-correspondence aggregation network for video
  super-resolution.
\newblock In {\em Proceedings of European Conference on Computer Vision
  (ECCV)}, 2020.

\bibitem{li2018learning}
Yue Li, Dong Liu, Houqiang Li, Li Li, Zhu Li, and Feng Wu.
\newblock Learning a convolutional neural network for image compact-resolution.
\newblock {\em IEEE Transactions on Image Processing (TIP)}, 2018.

\bibitem{lim2017enhanced}
Bee Lim, Sanghyun Son, Heewon Kim, Seungjun Nah, and Kyoung Mu~Lee.
\newblock Enhanced deep residual networks for single image super-resolution.
\newblock In {\em Proceedings of the IEEE Conference on Computer Vision and
  Pattern Recognition Workshops}, 2017.

\bibitem{6549107}
Ce Liu and Deqing Sun.
\newblock On bayesian adaptive video super resolution.
\newblock {\em IEEE Transactions on Pattern Analysis and Machine Intelligence
  (TPAMI)}, 2013.

\bibitem{7986143}
R. {Rassool}.
\newblock Vmaf reproducibility: Validating a perceptual practical video quality
  metric.
\newblock In {\em 2017 IEEE International Symposium on Broadband Multimedia
  Systems and Broadcasting (BMSB)}, 2017.

\bibitem{sajjadi2018frame}
Mehdi~SM Sajjadi, Raviteja Vemulapalli, and Matthew Brown.
\newblock Frame-recurrent video super-resolution.
\newblock In {\em Proceedings of the IEEE Conference on Computer Vision and
  Pattern Recognition (CVPR)}, 2018.

\bibitem{shannon1949communication}
Claude~Elwood Shannon.
\newblock Communication in the presence of noise.
\newblock {\em Proceedings of the IRE}, 1949.

\bibitem{sun2020learned}
Wanjie Sun and Zhenzhong Chen.
\newblock Learned image downscaling for upscaling using content adaptive
  resampler.
\newblock {\em IEEE Transactions on Image Processing (TIP)}, 2020.

\bibitem{tao2017detail}
Xin Tao, Hongyun Gao, Renjie Liao, Jue Wang, and Jiaya Jia.
\newblock Detail-revealing deep video super-resolution.
\newblock In {\em Proceedings of the IEEE Conference on Computer Vision and
  Pattern Recognition (CVPR)}, 2017.

\bibitem{tian2020tdan}
Yapeng Tian, Yulun Zhang, Yun Fu, and Chenliang Xu.
\newblock Tdan: Temporally-deformable alignment network for video
  super-resolution.
\newblock In {\em Proceedings of the IEEE Conference on Computer Vision and
  Pattern Recognition (CVPR)}, 2020.

\bibitem{wang2019edvr}
Xintao Wang, Kelvin~CK Chan, Ke Yu, Chao Dong, and Chen Change~Loy.
\newblock Edvr: Video restoration with enhanced deformable convolutional
  networks.
\newblock In {\em Proceedings of the IEEE Conference on Computer Vision and
  Pattern Recognition Workshops}, 2019.

\bibitem{wang2018esrgan}
Xintao Wang, Ke Yu, Shixiang Wu, Jinjin Gu, Yihao Liu, Chao Dong, Yu Qiao, and
  Chen~Change Loy.
\newblock Esrgan: Enhanced super-resolution generative adversarial networks.
\newblock In {\em The European Conference on Computer Vision Workshops}, 2018.

\bibitem{wang2017predrnn}
Yunbo Wang, Mingsheng Long, Jianmin Wang, Zhifeng Gao, and S~Yu Philip.
\newblock Predrnn: Recurrent neural networks for predictive learning using
  spatiotemporal lstms.
\newblock In {\em Advances in Neural Information Processing Systems (NIPS)},
  2017.

\bibitem{wang2004image}
Zhou Wang, Alan~C Bovik, Hamid~R Sheikh, and Eero~P Simoncelli.
\newblock Image quality assessment: from error visibility to structural
  similarity.
\newblock {\em IEEE Transactions on Image Processing (TIP)}, 2004.

\bibitem{xiao2020invertible}
Mingqing Xiao, Shuxin Zheng, Chang Liu, Yaolong Wang, Di He, Guolin Ke, Jiang
  Bian, Zhouchen Lin, and Tie-Yan Liu.
\newblock Invertible image rescaling.
\newblock In {\em Proceedings of European Conference on Computer Vision
  (ECCV)}, 2020.

\bibitem{Xue_2019}
Tianfan Xue, Baian Chen, Jiajun Wu, Donglai Wei, and William~T Freeman.
\newblock Video enhancement with task-oriented flow.
\newblock {\em International Journal of Computer Vision (IJCV)}, 2019.

\bibitem{yi2019progressive}
Peng Yi, Zhongyuan Wang, Kui Jiang, Junjun Jiang, and Jiayi Ma.
\newblock Progressive fusion video super-resolution network via exploiting
  non-local spatio-temporal correlations.
\newblock In {\em Proceedings of IEEE International Conference on Computer
  Vision (ICCV)}, 2019.

\bibitem{zhang2018image}
Yulun Zhang, Kunpeng Li, Kai Li, Lichen Wang, Bineng Zhong, and Yun Fu.
\newblock Image super-resolution using very deep residual channel attention
  networks.
\newblock In {\em Proceedings of European Conference on Computer Vision
  (ECCV)}, 2018.

\end{thebibliography}
